\documentclass[final]{cvpr}
\usepackage{times}
\usepackage{epsfig}
\usepackage{graphicx}
\usepackage{amsmath}
\usepackage{amssymb}
\usepackage{booktabs}

\usepackage{color}
\usepackage[pagebackref=true,breaklinks=true,colorlinks,bookmarks=false]{hyperref}

\def\cvprPaperID{2080} 
\def\confYear{CVPR 2021}

\begin{document}

\title{DANNet: A One-Stage Domain Adaptation Network for Unsupervised \\ Nighttime Semantic Segmentation}

\author{Xinyi Wu$^{1*}$, Zhenyao Wu$^{1,2}$\thanks{Equal contribution.}, Hao Guo$^{1}$, Lili Ju$^{1\dag}$, Song Wang$^{1}$\thanks{Co-corresponding authors. Code is available at \url{https://github.com/W-zx-Y/DANNet}.}\\
	$^1$University of South Carolina, USA \qquad\qquad
    $^2$Farsee2 Technology Ltd, China\\
	{\tt\small \{xinyiw, zhenyao, hguo\}@email.sc.edu,  ju@math.sc.edu, songwang@cec.sc.edu}
}
\maketitle
\pagestyle{empty}
\thispagestyle{empty}

\begin{abstract}
	Semantic segmentation of nighttime images plays an equally important role as that of daytime images in autonomous driving,
	but the former is much more challenging due to poor illuminations and arduous human annotations. In this paper, we propose a novel domain adaptation network (DANNet) for nighttime semantic segmentation without using labeled nighttime image data. It employs an adversarial training with a labeled daytime dataset and an unlabeled dataset that contains coarsely aligned day-night image pairs. Specifically, for the unlabeled day-night image pairs, we use the pixel-level predictions of static object categories on a daytime image as a pseudo supervision to segment its counterpart nighttime image. We further design a re-weighting strategy to handle the inaccuracy caused by misalignment between day-night image pairs and wrong predictions of daytime images, as well as boost the prediction accuracy of small objects. The proposed DANNet is the first one-stage adaptation framework for nighttime semantic segmentation, which does not train additional day-night image transfer models as a separate pre-processing stage. Extensive experiments on Dark Zurich and Nighttime Driving datasets show that our method achieves state-of-the-art performance for nighttime semantic segmentation.
\end{abstract}

\section{Introduction}

\begin{figure}[htbp]
	\begin{center}
		\begin{tabular}{ccccc}
			\hspace{-.11cm}\includegraphics[width=.224\textwidth]{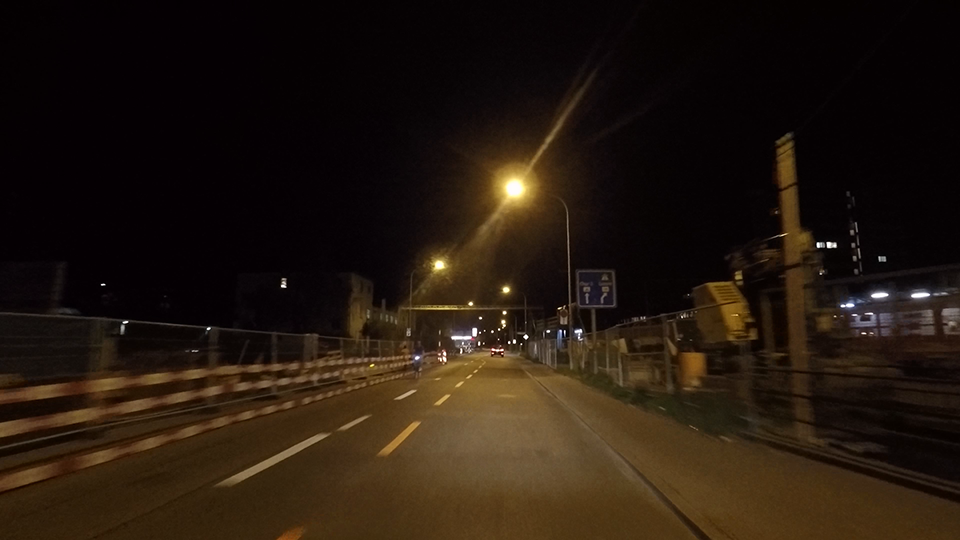} & \hspace{-.5cm}
			\includegraphics[width=.224\textwidth]{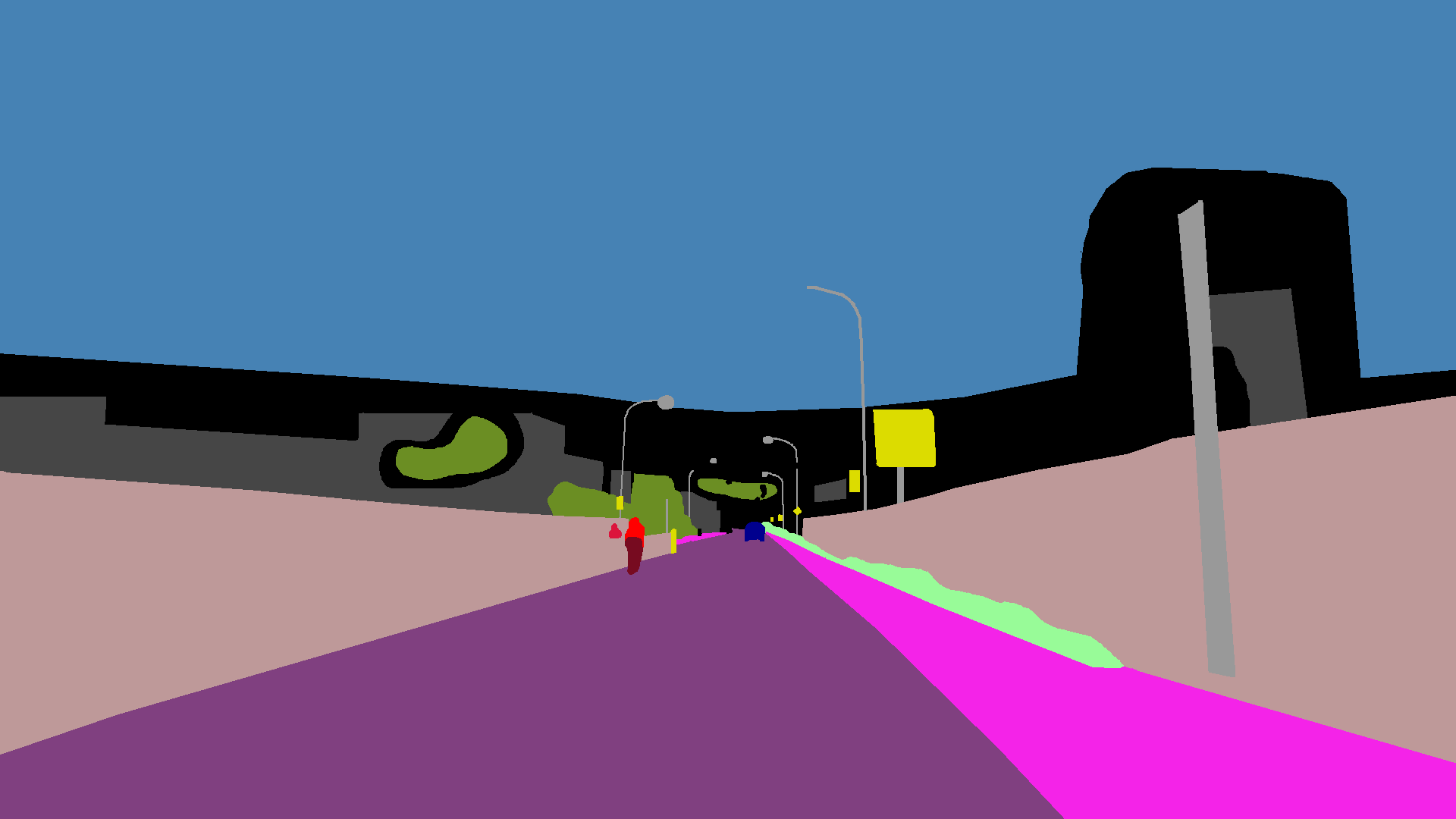}\vspace{-0.1cm} \\
			\hspace{-.11cm} \small Input image & \hspace{-.5cm} \small Ground truth\\
			\hspace{-.11cm}\includegraphics[width=.224\textwidth]{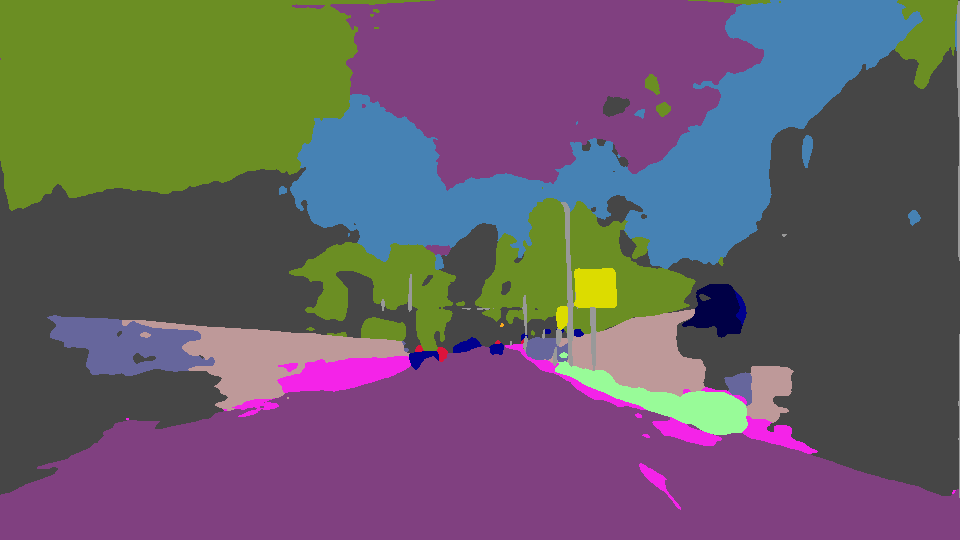} & \hspace{-.4cm}
			\includegraphics[width=.224\textwidth]{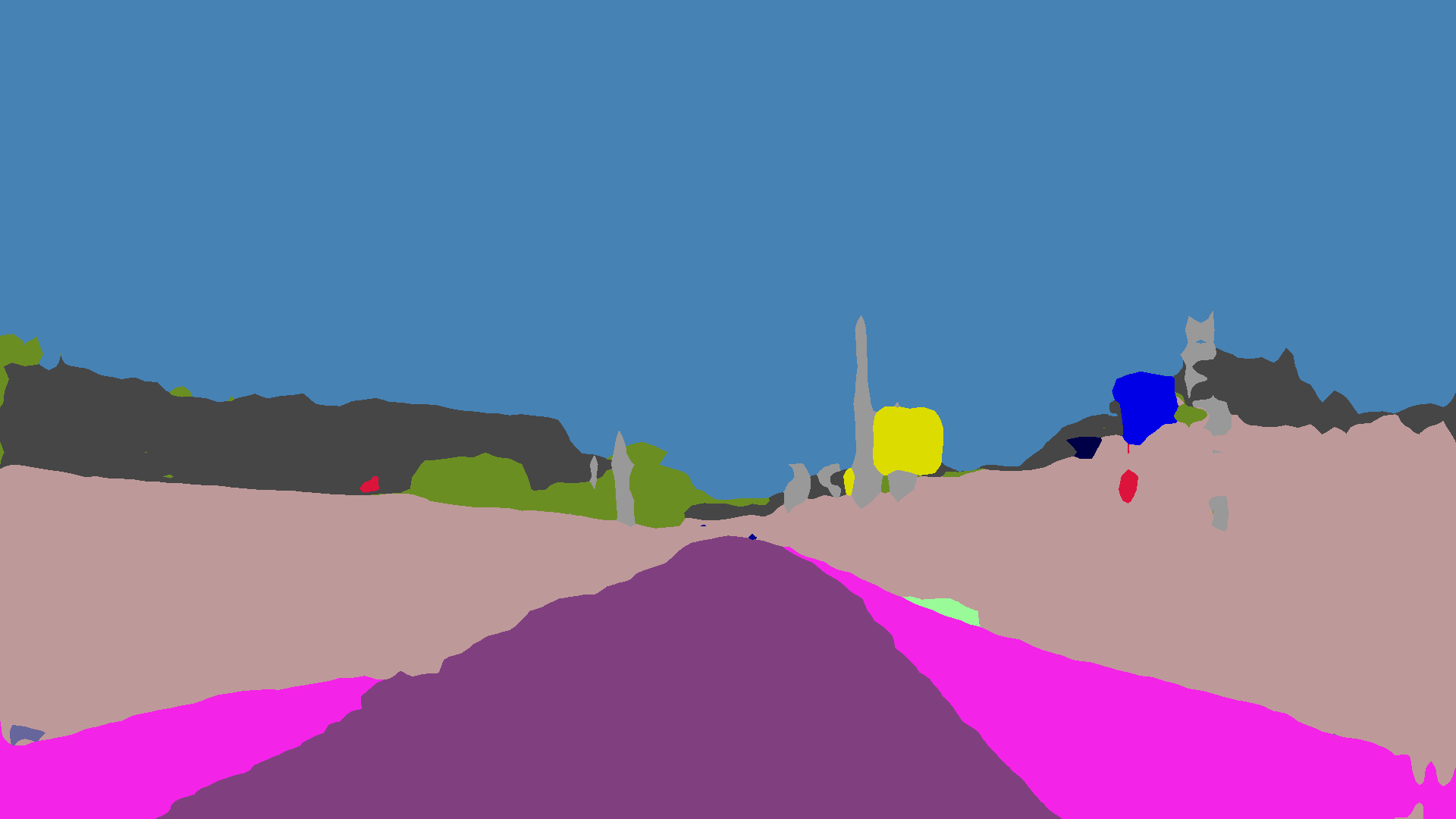} \vspace{-0.1cm}\\
			\hspace{-.11cm} \small MGCDA & \hspace{-.5cm} \small DANNet (Ours)\\
		\end{tabular}
		\vspace{-4pt}
		\caption{Visual comparison of the nighttime semantic segmentation results between the state-of-the-art transfer-based approach ``MGCDA'' \cite{sakaridis2020map} and our proposed DANNet.}
		\label{introduction}
		\vspace{-15pt}
	\end{center}
\end{figure}

Aiming to label each pixel of a given image to an object category, semantic segmentation is a fundamental computer vision task and plays an important role in many applications such as autonomous driving~\cite{geigerwe}, medical imaging~\cite{chen2019learning} and human parsing~\cite{zhang2020part}. 
With the advancement of deep learning and computing power, the state-of-the-art performance of semantic segmentation for natural scene images taken at the daytime has been significantly  improved in recent years~\cite{fu2019dual,huang2019ccnet}. 
Many researchers have started to segment more challenging images under various kinds of degradations, such as those taken in foggy weather~\cite{sakaridis2018semantic} or at the nighttime~\cite{sakaridis2019guided}.
In this paper, we focus on semantic segmentation of nighttime images, which has wide and important applications in autonomous driving.

With many indiscernible regions and visual hazards~\cite{zendel2017good}, e.g., under/over exposure and motion blur, it is usually difficult even for human to build high-quality pixel-level annotations of the nighttime scene images as ground truth, which, however, is a prerequisite for training many deep neural networks for semantic image segmentation.  
To handle this problem, several domain adaptation methods have been proposed to transfer the semantic segmentation models from daytime to nighttime without using labels in the nighttime domain. 
For example, in~\cite{dai2018dark,sakaridis2019guided,sakaridis2020map}, an intermediate twilight domain is taken as a bridge to build the adaptation between daytime to nighttime.
In~\cite{sakaridis2019guided,romera2019bridging,sun2019see,nag2019s,sakaridis2020map}, an image transferring network is trained to stylize nighttime or daytime images and construct synthetic datasets. 
All these methods require an additional pre-processing stage of training an image transfer model between daytime and nighttime.
This is not only time-consuming but also making the second stage closely rely on the first one. Especially, it is difficult to generate a transferred image that shares the exactly same semantic information with the original images when the domain gap is large.

In this paper, we propose a novel one-stage domain adaptation network (DANNet) based on adversarial learning for nighttime semantic segmentation (shown in Figure~\ref{introduction}) by using the newly released Dark Zurich dataset~\cite{sakaridis2019guided},  which contains unlabeled day-night scene image pairs that are coarsely aligned using GPS recordings.
The proposed DANNet performs a multi-target adaptation from Cityscapes data to Dark Zurich daytime  (Dark Zurich-D) and nighttime data (Dark Zurich-N). 
Specifically, we first adapt the model from Cityscapes, which contains large-scale training data with labels, to Dark Zurich-D since they are all taken at the daytime. 
Then, the prediction of Dark Zurich-D is used as a pseudo supervision for Dark Zurich-N in the network training. 
We apply an image  relighting subnetwork to make the intensity distribution of the images from different domains to be close. 
Following~\cite{tsai2018learning}, we incorporate a weight-sharing semantic segmentation network to make predictions for the relighted images and 
perform an adversarial learning in the output space to ensure very close layout across different domains. 
We further design a re-weighting strategy to handle the inaccuracy caused by misalignment between day-night image pairs and wrong predictions of daytime images, as well as 
boost the prediction accuracy of small objects. 
We conduct extensive experiments on Dark Zurich and Nighttime Driving datasets to justify the effectiveness of the proposed DANNet for nighttime semantic segmentation. The main contributions of our work are summarized in the following:

\begin{itemize}
	\vspace{-3pt}
	\item We propose a multi-target domain adaptation network, DANNet, for nighttime semantic segmentation via adversarial learning. DANNet consists of an image relighting network and a semantic segmentation network, as well as two discriminators. To the best of our knowledge, the proposed DANNet is the first one-stage adaptation framework for nighttime semantic segmentation. 
	\item We demonstrate that the segmentation of Dark Zurich-D images can provide pseudo supervision for segmenting the corresponding Dark Zurich-N images, by considering only static object categories. In particular, it is  shown that the specially designed probability re-weighting strategy can significantly enhance the segmentation of small objects.
	\item Experiments on Dark Zurich-test and Nighttime Driving datasets show that the proposed DANNet achieves a new state-of-the-art performance of nighttime semantic segmentation. Ablation study also verifies the effectiveness of each component in DANNet.
\end{itemize}

\begin{figure*}[htbp]
	\centering{
		\resizebox{17.2cm}{!}{\includegraphics{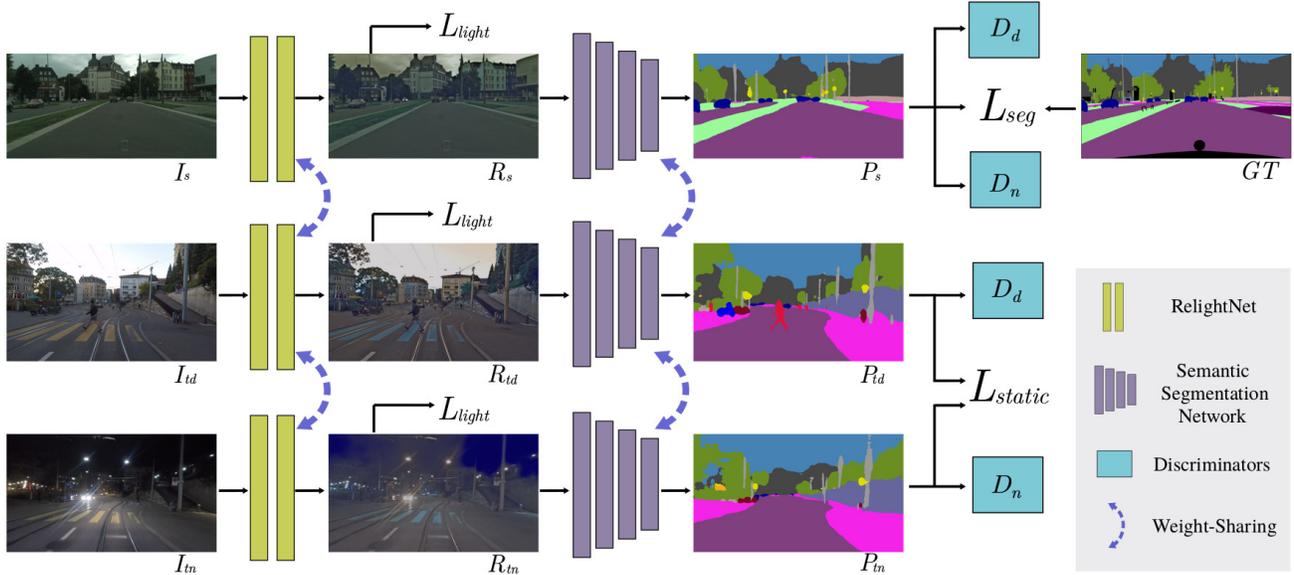}}
		\vspace{3pt}
		\caption{The architecture  of the proposed DANNet. Three input images $I_s$, $I_{td}$, and $I_{tn}$ are from the source domain $S$ (Cityscapes) and two target domains  $T_d$ and $T_n$ (Dark Zurich-D and Dark Zurich-N), respectively.
			They go through a weight-sharing image relighting network which can make their distributions to be close to each other using the light loss $L_{light}$. 
			All the outputs are fed into a weight-sharing segmentation network to obtain the predictions. 
			For the predictions from $I_s$, a semantic segmentation loss $L_{seg}$ is computed using the ground truth from the source dataset. 
			Besides, the predictions from $I_{td}$ for the categories of static objects provide weak supervision for the corresponding categories from $I_{tn}$, reflected by a static loss $L_{static}$. Note that the composition of the relighting network and the semantic segmentation network forms the {generator} $G$. 
			Two discriminators $D_d$ and $D_n$ are proposed to distinguish outputs from the source domain $S$  or the target domains $T_d$  and from the source domain $S$  or the  target domains $T_n$, respectively.}
		\label{network}
	}\vspace{-2pt}
\end{figure*}
	
\section{Related Work}

\noindent{\bf Domain adaptation for semantic segmentation}\hspace{0.2cm}
Domain adaptation methods are developed to transfer knowledge learned from source domains to target domains which share similar objects yet different data distributions. 
Recently, domain adaptation has been applied to help semantic segmentation. 
In~\cite{hoffman2016fcns}, Hoffman \etal proposed a novel fully convolutional domain adversarial learning approach with category constraints~\cite{pathak2015constrained} for semantic segmentation. 
Tsai \etal~\cite{tsai2018learning} later developed a multi-level adversarial network to perform domain adaptation in the output space.  

Instead of using adversarial learning techniques, image translation and style transfer~\cite{zhu2017unpaired} from source images to target ones, or vice versa, have been widely used for domain adaptation~\cite{hoffman2018cycada,wu2018dcan}. 
Previous works have shown that domain-invariant representations can be obtained in the process of image translation between the source and target domains~\cite{sankaranarayanan2018learning,zhu2018penalizing,chang2019all}. 
Several recent works~\cite{li2019bidirectional,wang2020differential,kim2020learning} made use of self-training strategies by iteratively predicting and fine-tuning a set of pseudo labels in multiple rounds of network training.
Another line of researches~\cite{zhang2017curriculum,lian2019constructing} adopted the curriculum-style learning by first learning easy properties in the target domain and then using it to regularize the semantic segmentation model.
However, most of these general-purpose domain adaptation approaches cannot handle well the significant adaptation gap between the daytime and the nighttime images and therefore could not achieve satisfactory performance in nighttime semantic segmentation~\cite{sakaridis2019guided}.
Specifically, all the above methods focus on the domain adaptation for synthetic-to-real (i.e., GTA5 \cite{richter2016playing} or SYNTHIA \cite{ros2016synthia} to Cityscapes) or cross-city images (i.e., Cityscapes to Cross-City \cite{chen2017no}), which are all daytime to daytime adaptations.
In this paper, we instead focus on the adaptation between the daytime and the nighttime domains with significantly different illumination patterns~\cite{sakaridis2019guided}.
 
\vspace{0.1cm}
\noindent{\bf Nighttime semantic segmentation}\hspace{0.2cm}
Recently, Dai \etal~\cite{dai2018dark} leveraged an intermediate twilight domain to progressively adapt semantic models trained in daytime scenes to nighttime. 
Sakaridis \etal~\cite{sakaridis2019guided,sakaridis2020map} further extended it to a guided curriculum adaptation framework, which uses both the stylized synthetic images and the unlabeled real images to exploit the cross-time-of-day correspondence of the scene images. 
However, such gradual adaptation approaches usually need to train multiple semantic segmentation models, e.g., three models in~\cite{sakaridis2019guided} for three different domains respectively, 
which is highly inefficient.  
Following works along this line~\cite{romera2019bridging,sun2019see,nag2019s} also train some additional image transfer models, e.g., CycleGAN \cite{zhu2017unpaired}, to perform the day-to-night or night-to-day image transfer before training the semantic segmentation models. 
For these methods, the performance of later adaptation and semantic segmentation is highly dependent on the image transfer model pre-trained in the pre-processing stage.

Vertens \etal~\cite{vertens2020heatnet} proposed to leverage the thermal infrared images as a complementary input to the RGB images for nighttime semantic segmentation since thermal radiation is not very sensitive to the illumination changes. 
In~\cite{di2020rainy}, a two-stage adversarial training method was proposed  for  semantic segmentation of rainy night scenes by performing domain adaptation between day-night near scene pairs. 
Different from all the above methods, the DANNet proposed  in this paper performs a one-stage end-to-end adversarial learning for training the nighttime semantic segmentation network without using any other image modalities.

\section{Proposed Method}
\subsection{Framework overview}
Our method involves a source domain $S$ and two target domains $T_d$ and $T_n$, where $S$, $T_d$, and $T_n$ represent Cityscapes (daytime), Dark Zurich-D (daytime), and Dark Zurich-N (nighttime), respectively. 
Note that only the source domain $S$ of Cityscapes has ground-truth semantic segmentation in training. 
The proposed DANNet proceeds the domain adaptation from $S$ to $T_d$ and $S$ to $T_n$ simultaneously and it consists of three different modules: an 
image relighting network, a semantic segmentation network, and two discriminators, as illustrated in Figure~\ref{network}.

\subsection{Network architecture}
\begin{figure*}[htbp]
	\centering{
		\resizebox{17.2cm}{!}{\includegraphics{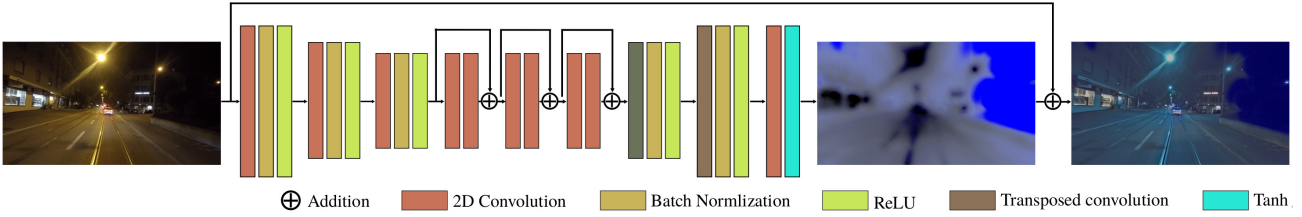}}
		\vspace{3pt}
		\caption{The  structure of the image relighting network. 
			It consists of four convolutional layers, three residual blocks and two transposed convolutional layers, and each convolutional layer is followed by a batch normalization layer. 
			The output from the last layer is then added to the input images to obtain the relighted image.}
		\label{relight}
	}\vspace{-4pt}
\end{figure*}
All modules of the proposed DANNet are elaborated in detail below.

\vspace{0.1cm}
\noindent{\bf Image relighting network}
Inspired by~\cite{jenicek2019no}, we design an image  relighting  network to make the intensity distributions of the images from different domains to be close such that the later semantic segmentation network is less sensitive to illumination changes. 
The relighting network takes the scene images $I_s$, $I_{td}$ and $I_{tn}$  from the three domains, and generates the relighted images $R_s$, $R_{td}$ and $R_{tn}$, respectively.
The relighting network shares weights for all input images from the three domains, see Figure \ref{relight} for the detailed  structure of this network. 

\vspace{0.1cm}
\noindent{\bf Semantic segmentation network}\hspace{0.2cm}
We select and test three popular semantic segmentation networks in our method: Deeplab-v2~\cite{chen2017deeplab}, RefineNet~\cite{Lin:2017:RefineNet} and PSPNet~\cite{zhao2017pyramid}. Note that the
common backbone is ResNet-101~\cite{he2016deep}  in all of them.
For this module, we share weights for all the input images from the three domains. 
The semantic segmentation network takes $R_s$, $R_{td}$ and $R_{tn}$ as the inputs and produces segmentation predictions (category-likelihood map) $P_s$, $P_{td}$ and $P_{tn}$ for the three domains, respectively. The composition of
the image relighting network and the semantic segmentation network forms the {generator} $G$ of the proposed DANNet.

\vspace{0.1cm}
\noindent{\bf Discriminators}\hspace{0.2cm}
As done in~\cite{tsai2018learning}, the discriminators are designed to distinguish whether the segmentation prediction comes from the source domain or either of the target domains by performing adversarial learning in the output space. 
We modified the architecture in~\cite{radford2015unsupervised} following \cite{tsai2018learning} by utilizing all fully convolutional layers. 
Particularly, it includes 5 convolutional layers with the channel numbers of $\{64, 128, 256, 256, 1\}$, and a kernel size of $4 \times 4$. 
The stride is 2 for the first two convolutional layers and 1 for the rest.
Since we have two target domains $T_d$ and $T_n$, we design two discriminators $D_d$ and $D_n$ to distinguish whether the output is from $S$ or $T_d$ and from $S$ or $T_n$, respectively.
The two discriminators share the same structures yet the weights and are jointly trained.  

\subsection{Probability re-weighting} \hspace{0.2cm}
Due to the fact that the numbers of pixels for different object categories are imbalanced in the source domain, 
network training can usually converge more easily by predicting a pixel to be a category of large-size object, such as road, building, and tree, in training discriminators.
In this case, it is quite difficult to correctly predict the pixels of small objects which have relatively fewer annotations in the dataset, such as pole, sign, and light. 
To address this problem, we propose a re-weighting strategy to the predicted category-likelihood maps.
Specifically, for each category $k\in \mathbb{C}$, we first define a weight
\begin{equation}
w'_k = -\log(a_k),
\end{equation}
where $a_k$ is the proportion of all the valid pixels that are labeled as category $k$ in the source domain. 
Clearly the smaller the value of $a_k$, the larger the value of $w'_k$ and the use of such a weight can help segment the categories of smaller-size objects.
We use the logarithm to prevent from overweighting small-size object categories.
In our experiment, we further normalize this weight by
\begin{equation}\label{pr}
w_k = \frac{w'_k -\overline{w}}{\sigma(w)}\cdot std+ avg,
\end{equation}
where $\overline{w}$ and $\sigma(w)$ are the mean and standard deviation of $w'_k, k\in \mathbb{C}$, respectively. The parameters 
$std$ and $avg$ are two positive constants we pre-select to shift the value range of $w_k$ to be mainly positive.
During training, we set $std=0.05$ and $avg=1.0$ empirically. 
We then multiply each normalized weight $w_k$ with the corresponding category channel of the predicted likelihood map $P$, where $P \in \{P_{td}, P_{tn}\}$.
Thus, the final semantic segmentation result $F$ is obtained by employing an argmax operation on the multiplication result.

\subsection{Objective functions}
In this subsection, we introduce all the objective functions involved in the proposed end-to-end DANNet training, including the light loss, the semantic segmentation loss, the static loss, and the adversarial loss.

\vspace{0.1cm}
\noindent{\bf Light Loss}\hspace{0.2cm}
The light loss is proposed to ensure that the intensity distributions of the outputs $R_s$, $R_{td}$ and $R_{tn}$ after the image relighting network are close to each other. 
The light loss is a combination of three loss functions: the total variation loss $L_{tv}$, the exposure control loss $L_{exp}$, and the structural similarity loss $L_{ssim}$.

The total variation loss $L_{tv}$~\cite{rudin1992nonlinear} is widely used in image denoising~\cite{zhang2017beyond} and image synthesis~\cite{wang2018high} to make images smoother. 
In this paper, we apply such a loss function to remove rough textures such as noises to facilitate the semantic segmentation. The loss $L_{tv}$ is defined by
\begin{equation}
L_{tv} = \frac{1}{N}\|(\nabla_x(I-R))^2+(\nabla_y(I-R))^2\|_1,
\end{equation}
where $I \in \{I_s, I_{td}, I_{tn}\}$ represents the input images, $R \in \{R_s, R_{td}, R_{tn}\}$ is the output of the relighting network, $N$ is the number of pixels in $I$, $\nabla_x$ and $\nabla_y$ represent intensity gradients between neighboring pixels along the $x$ and $y$ directions, respectively, and $\|\cdot \|_1$ is the $L_1$ norm that sums up over all the pixels. 

To obtain the similar lighting effects in the day and night scenarios, we apply the following exposure loss $L_{exp}$ proposed in~\cite{Zero-DCE} to control the exposure level:
\begin{equation}
L_{exp} = \frac{1}{M}\|\varphi (R)-E\|_1,
\end{equation}
where $\varphi$ is a $32 \times 32$ average pooling function and $M$ represents the number of pixels in $\varphi (R)$. 
Different from~\cite{Zero-DCE}, the value of $E$ is dynamically set to be the average intensity value of the nighttime image for each training iteration. 

The structural similarity loss $L_{ssim}$~\cite{wang2004image} is widely used for image reconstruction~\cite{godard2017unsupervised,chen2018learning}.
Here we apply this loss function to ensure that the generated relighted images $R$ could maintain the structure of original images $I$. The loss $L_{ssim}$ is defined by
\begin{equation}
L_{ssim} = \frac{1}{2N}\|1-SSIM(I, R)\|_1.
\end{equation}
As in~\cite{godard2017unsupervised}, we use a simplified SSIM (structural similarity index measure)  with a $3 \times 3$ block filter in this loss function.

Finally, by combining all the three loss terms, our light loss $L_{light}$ is defined by
\begin{equation}
L_{light} = \alpha_{tv}L_{tv} + \alpha_{exp}L_{exp} +\alpha_{ssim}L_{ssim},
\end{equation}
where $\alpha_{tv},  \alpha_{exp}$, and $\alpha_{ssim}$ are set to 10, 1, and 1, respectively in all experiments.

\vspace{0.1cm}
\noindent{\bf Semantic segmentation loss}\hspace{0.2cm}
We adopt the widely used weighted cross-entropy loss for training the semantic image segmentation in the source domain:
\begin{equation}
L_{seg} = -\frac{1}{N |\mathbb{C}|}\sum _{k\in \mathbb{C}}\|w_k GT^{(k)} \cdot log(P_s^{(k)}) \|_1,
\end{equation}
where $P_s^{(k)}$ is the $k$-th channel of the prediction $P_s$ from the source images, $w_k$ is the weight defined in Eq.~(\ref{pr}), and $GT^{(k)}$ is the one-hot encoding of the ground truth for the $k$-th category.

\vspace{0.1cm}
\noindent{\bf Static loss}\hspace{0.2cm}
Based on the fact that the daytime image share similarities with its corresponding nighttime counterpart when considering only the static object categories, 
we here introduce a static loss to provide  pixel-level pseudo supervision for the static object categories, e.g., road, sidewalk, wall, fence, pole, light, sign, vegetation, terrain and sky, in the nighttime images. 

Given the segmentation predictions $P_{td} \in \mathbb{R}^{H \times W \times C}$ and $P_{tn} \in \mathbb{R}^{H \times W \times C} $, we only consider the channels corresponding to the static categories 
for calculating this loss. 
Let us denote $C^\mathcal S$ as the total number of the categories of static objects, then it holds that
 $P_{td}^\mathcal S \in \mathbb{R}^{H \times W \times C^ \mathcal S}$ and $P_{tn}^\mathcal S \in \mathbb{R}^{H \times W \times C^ \mathcal S}$.

We first apply Eq.~(\ref{pr}) to calculate the re-weighted prediction $F_{td}$ as the pseudo label. 
Following~\cite{yu2018learning,chen2019domain}, we then  employ the focal loss~\cite{lin2017focal} to remedy the imbalance among different categories of training samples. 
Finally, the static loss $L_{static}$ is defined by
\begin{equation}
L_{static} = -\frac{1}{N}\|(1 - P_{tn}^\mathcal S)^\gamma  log(p)\|_1, 
\end{equation}
where $N$ is the total number of valid pixels in the segmentation ground truth,  $\gamma$ is the focusing parameter (set to 1 in all experiments), and $p$ is the likelihood map for the correct category.
Different from the focal loss in~\cite{lin2017focal}, we compute $p$ at each pixel $i$ in a $3 \times 3$ local region for category $c$ by
\begin{equation}
p(c,i) = \max_{j}( o(c,j) \cdot P_{tn}^\mathcal S(c,i)),
\end{equation}
where $o$ is the one-hot encoding of the semantic pseudo ground truth $F_{td}$, and $j$ represents each position of the $3 \times 3$ region centered at $i$.

\vspace{0.1cm}
\noindent{\bf Adversarial loss}\hspace{0.2cm}
We employ two discriminators for adversarial learning, which are used to distinguish whether the output is from the source domain or one of the two target domains, i.e., $S$ or $T_d$ and $S$ or $T_n$. 
We adopt the least-squares loss function~\cite{mao2017least} to make both predictions $P_{td}$ and $P_{tn}$ to be close to $P_s$. 
Specifically, we define the combination of these two adversarial losses ($L_{adv}$) as:
\begin{equation}
 L_{adv} = (D_{d}(P_{td})-r)^2 + (D_{n}(P_{tn})-r)^2,
\end{equation}
where $P_{td}=G(I_{td})$, $P_{tn}=G(I_{tn})$, and $r$ is the label for the source domain which has the same resolution as the output of discriminators.  
Thus, the total loss $L_{total}$ of the generator (G) is defined by combining $L_{light} $, $L_{seg}$, $L_{static}$ and $L_{adv}$:
\begin{equation}\small
\min\limits_G L_{total} = \beta_1 L_{light}+ \beta_2 L_{seg} + \beta_3 L_{static} +\beta_4 L_{adv},
\end{equation}
where $\beta_1,  \beta_2, \beta_3$, and $\beta_4$ are set to 0.01, 1, 1 and 0.01 respectively in all experiments.

The generator and the corresponding discriminators are trained alternatively and the objective functions of the discriminators $D_s$ and $D_n$ are defined respectively by:
\begin{equation}
\begin{aligned}
\min\limits_{D_{d}}L_{d}=  \frac{1}{2} (D_{d}(P_s)-r)^2+ 
 \frac{1}{2} (D_{d}(P_{td})-f)^2,   
\end{aligned}
\end{equation}
\vspace{-0.2cm}
\begin{equation}
\begin{aligned}
\min\limits_{D_{n}}L_{n} =  \frac{1}{2} (D_{n}(P_s)-r)^2 +  \frac{1}{2} (D_{n}(P_{tn})-f)^2,   
\end{aligned}
\end{equation}
where $f$ is the label for the target domains with the same resolution as the output of discriminators.

\section{Experiments}

\subsection{Datasets and evaluation metrics}

For all experiments, we use the mean of category-wise intersection-over-union (mIoU) as the evaluation metric, and the higher the better.
The following datasets are used for model training and performance evaluation:

\vspace{0.1cm}
\noindent{\bf Cityscapes} \cite{cordts2016cityscapes}\hspace{0.2cm}
The Cityscapes dataset contains 5,000 frames taken in street scenes with pixel-level annotations of a total of 19 categories, and both the original images and annotations have a resolution of 
$2,048 \times 1,024$ pixels. 
In total, there are 2,975 images for training, 500 images for validation and 1,525 images for testing. In this paper,
we use the Cityscapes training set in the training stage of the proposed DANNet for adversarial learning.

\vspace{0.1cm}
\noindent{\bf Dark Zurich} \cite{sakaridis2019guided}\hspace{0.2cm}
The Dark Zurich dataset consists of 2,416 nighttime images, 2,920 twilight images and 3,041 daytime images for training, which are all unlabeled with a resolution of $1,920 \times 1,080$. 
Images in these three domains can be coarsely aligned by using GPS-based nearest neighbor assignment to compensate the translation in each direction and the zoom in/out factors. 
In this paper, we only use 2,416 night-day image pairs in training of the proposed DANNet (without using the twilight images). 
The Dark Zurich dataset also contains another 201 annotated nighttime images including 50 for validation (Dark Zurich-val) and 151 for testing (Dark Zurich-test), for quantitative evaluation.
Note that the Dark Zurich-test serves as an online benchmark whose ground truth are not publicly available.
In our experiments, by submitting the segmentation results to the online evaluation website we get the performance of the proposed DANNet on Dark Zurich-test against the annotated ground truths. 

\vspace{0.1cm}
\noindent{\bf Nighttime Driving} \cite{dai2018dark}\hspace{0.2cm}
The Nighttime Driving test set contains 50  nighttime images of resolution $1,920 \times 1,080$ from diverse visual scenes.
All these 50 images have been annotated at the pixel level using the same 19 Cityscapes category labels.
In our experiments, we only use Nighttime Driving test set for method evaluation.

\vspace{0.1cm}
\subsection{Experimental settings}
We implement the proposed DANNet using PyTorch on a single Nvidia 2080Ti GPU. 
Following~\cite{chen2017deeplab}, we train our network using the Stochastic Gradient Descent (SGD) optimizer with a momentum of 0.9 and a weight decay of $5 \times 10^{-4}$. 
The base learning rate is set to $2.5 \times 10^{-4}$ and then we employ the poly learning rate policy to decrease it with a power of 0.9. 
The batch size is set to 2. 
We use Adam optimizer~\cite{kingma2014adam} for training the discriminators with $\beta$ being set to $(0.9,0.99)$. 
The learning rate of the discriminators is set to $2.5 \times 10^{-4}$ and follows the same decay strategy as for the generator. 
In addition, we apply random cropping with the crop size of 512 on the scale between 0.5 and 1.0 for Cityscapes dataset, 
 with the crop size of 960 on the scale between 0.9 and 1.1 on Dark Zurich dataset, and  random horizontal flipping in the training.
 To make the training easier to converge, we use the semantic segmentation models that are pre-trained on Cityscapes for 150,000 epochs and report the performance of different segmentation models
on the validation set of Cityscapes and Dark Zurich in Table~\ref{baseline-cityscapes-val}.

\begin{table}[ht]
	\centering
	\caption{The mIoU performance of the pre-trained semantic segmentation models on the validation set of Cityscapes and Dark Zurich.}
	\vspace{4pt}
	\label{baseline-cityscapes-val}
	\renewcommand\arraystretch{1.0}
	\setlength{\tabcolsep}{2mm}{
		\footnotesize
		\begin{tabular}{lcc}
			\toprule
			Method  &  Cityscapes-val  &  Dark Zurich-val\\ 
			\midrule
			RefineNet \cite{Lin:2017:RefineNet}  &65.20 &15.16\\
			DeepLab-v2  \cite{chen2017deeplab}  &65.67 &12.14\\
			PSPNet \cite{zhao2017pyramid}  &63.37 &12.28\\
			\bottomrule
	\end{tabular}}
	\vspace{-3pt}
\end{table}

\begin{table*}[!ht]
	\centering
	\caption{The per-category results on Dark Zurich-test by current state-of-the-art methods and our DANNet. Cityscapes$\rightarrow$DZ-night denotes the adaptation from  Cityscapes  to Dark Zurich-night. The best
		results are presented in {\bf bold}, with the second best results \underline{underlined}.}
	\vspace{4pt}
	\label{DZ-test}
	\renewcommand\arraystretch{1.1}
	\footnotesize
	\setlength{\tabcolsep}{0.66mm}{
		\begin{tabular}{lcccccccccccccccccccc}
			\toprule
			Method  & \rotatebox{90}{road} & \rotatebox{90}{sidewalk} & \rotatebox{90}{building} & \rotatebox{90}{wall} & \rotatebox{90}{fence} & \rotatebox{90}{pole} & \rotatebox{90}{traffic light \ } & \rotatebox{90}{traffic sign} & \rotatebox{90}{vegetation} & \rotatebox{90}{terrain} & \rotatebox{90}{sky} & \rotatebox{90}{person} & \rotatebox{90}{rider} & \rotatebox{90}{car} & \rotatebox{90}{truck} & \rotatebox{90}{bus} & \rotatebox{90}{train} & \rotatebox{90}{motorcycle} & \rotatebox{90}{bicycle}  & \bf mIoU\\ 
			\midrule
			RefineNet \cite{Lin:2017:RefineNet}-Cityscapes   &  68.8 & 23.2 & 46.8 & 20.8 & 12.6 & 29.8 & 30.4 & 26.9 & 43.1 & 14.3 & 0.3 & 36.9 & 49.7 & 63.6 & 6.8 & \underline{0.2} & 24.0 & 33.6 & 9.3 & 28.5\\
			DeepLab-v2  \cite{chen2017deeplab}-Cityscapes   &  79.0 & 21.8 & 53.0 & 13.3 & 11.2 & 22.5 & 20.2 & 22.1 & 43.5 & 10.4 & 18.0 & 37.4 & 33.8 & 64.1 & 6.4 & 0.0 & 52.3 & 30.4 & 7.4 & 28.8\\
			PSPNet \cite{zhao2017pyramid}-Cityscapes  & 78.2 & 19.0 &  51.2 &  15.5 &  10.6 &  30.3 & 28.9 &  22.0 &  56.7 &  13.3 &  20.8 &  38.2 &  21.8 &  52.1 &  1.6 &  0.0 & 53.2 &  23.2 &  10.7 & 28.8\\
			\midrule
			AdaptSegNet-Cityscapes$\rightarrow$DZ-night \cite{tsai2018learning}  &  86.1 & 44.2 & 55.1 & 22.2 & 4.8 & 21.1 & 5.6 & 16.7 & 37.2 & 8.4 & 1.2 & 35.9 & 26.7 & 68.2 & 45.1 & 0.0 & 50.1 & 33.9 & 15.6 & 30.4 \\
			ADVENT-Cityscapes$\rightarrow$DZ-night \cite{vu2019advent} & 85.8 & 37.9 & 55.5 & 27.7 & 14.5 & 23.1 & 14.0 & 21.1 & 32.1 & 8.7 & 2.0 & 39.9 & 16.6 & 64.0 & 13.8 & 0.0 & 58.8 & 28.5 & 20.7 & 29.7 \\
			BDL-Cityscapes$\rightarrow$DZ-night \cite{li2019bidirectional}  & 85.3 & 41.1 & 61.9 & 32.7 & 17.4 & 20.6 & 11.4 & 21.3 & 29.4 & 8.9 & 1.1 & 37.4 & 22.1 & 63.2 & 28.2 & 0.0 & 47.7 & {\bf 39.4} & 15.7 & 30.8 \\
			DMAda \cite{dai2018dark}  & 75.5 & 29.1 & 48.6 & 21.3 & 14.3 & 34.3 & 36.8 & 29.9 & 49.4 & 13.8 & 0.4 & 43.3 & \underline{50.2} & 69.4 & 18.4 & 0.0 & 27.6 & 34.9 & 11.9 & 32.1 \\
			GCMA \cite{sakaridis2019guided}  & 81.7 & 46.9 & 58.8 & 22.0 & 20.0 &\underline{41.2} & {\bf 40.5} & {\bf 41.6} & 64.8 & 31.0 & 32.1 & {\bf 53.5} & 47.5 & {\bf 75.5} & \underline{39.2} & 0.0 & 49.6 & 30.7 & 21.0 & 42.0 \\
			MGCDA \cite{sakaridis2020map} & 80.3 & 49.3 & 66.2 & 7.8 & 11.0 & {\bf 41.4} & \underline{38.9} & \underline{39.0} & 64.1 & 18.0 & 55.8 & \underline{52.1} & {\bf 53.5} & \underline{74.7} & {\bf 66.0} & 0.0 & 37.5 & 29.1 & 22.7 & 42.5 \\
			\midrule
			DANNet (DeepLab-v2) & 88.6 & 53.4 & 69.8 & \underline{34.0} & 20.0 & 25.0 & 31.5 & 35.9 &69.5 & {\bf 32.2} & \underline{82.3} & 44.2 & 43.7 & 54.1 & 22.0 & 0.1 & 40.9 & 36.0 & {\bf 24.1} & 42.5\\
			DANNet (RefineNet) & \underline{90.0} & \underline{54.0} & {\bf 74.8} & {\bf 41.0} & \underline{21.1} & 25.0 & 26.8 & 30.2 & {\bf 72.0} 
			& 26.2 & {\bf 84.0} & 47.0 & 33.9 & 68.2 & 19.0 & {\bf 0.3} & \underline{66.4} &\underline{ 38.3} & \underline{23.6} & \underline{44.3}\\
			DANNet (PSPNet) & {\bf 90.4} & {\bf 60.1} & \underline{71.0} & 33.6 & {\bf 22.9} & 30.6 & 34.3 & 33.7 & \underline{70.5} & \underline{31.8} & 80.2 & 45.7 & 41.6 & 67.4 & 16.8 & 0.0 & {\bf 73.0} & 31.6 & 22.9 & {\bf 45.2}\\
			\bottomrule
	\end{tabular}}
\end{table*}

\begin{figure*}[h]\small
	\begin{center}
		\begin{tabular}{ccccc}
			\hspace{-.21cm}
			\includegraphics[width=.196\textwidth]{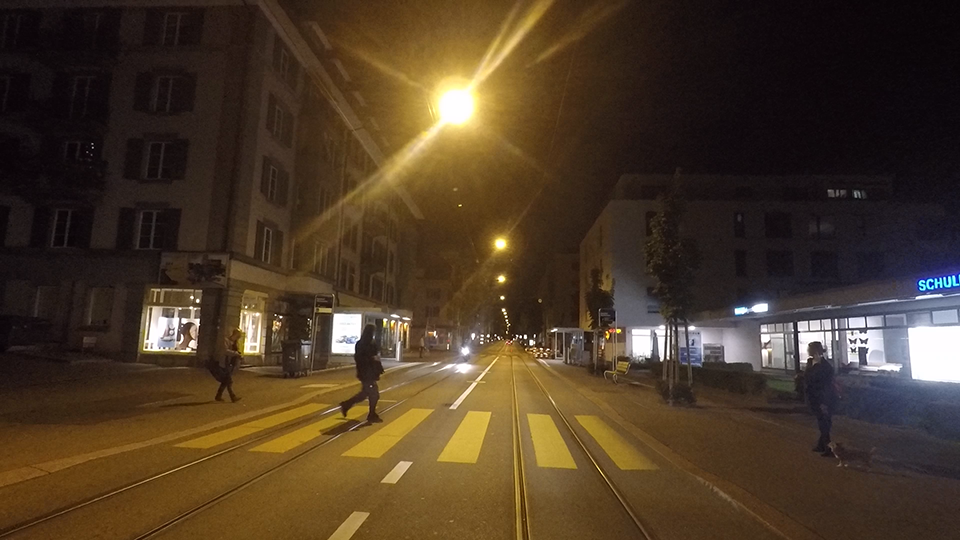} & \hspace{-.45cm}
			\includegraphics[width=.196\textwidth]{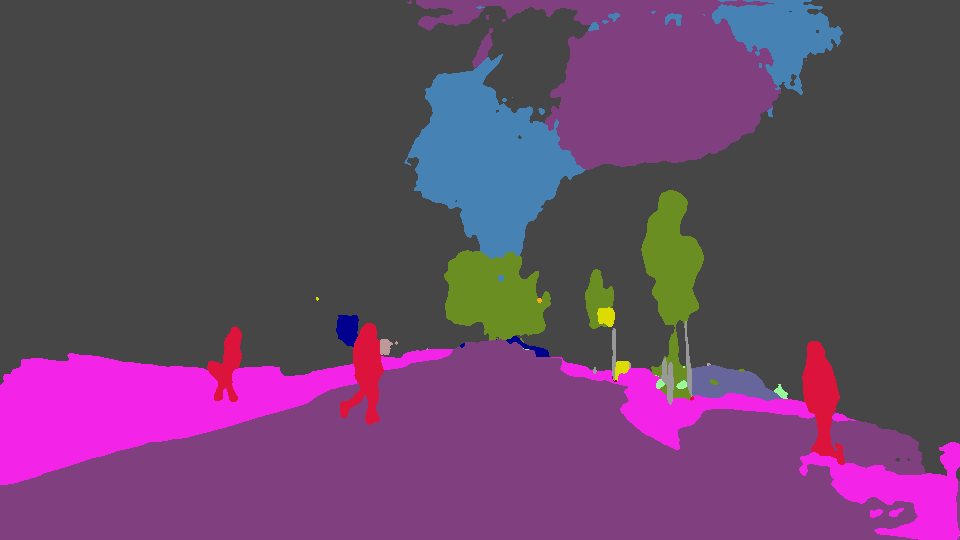} & \hspace{-.45cm}
			\includegraphics[width=.196\textwidth]{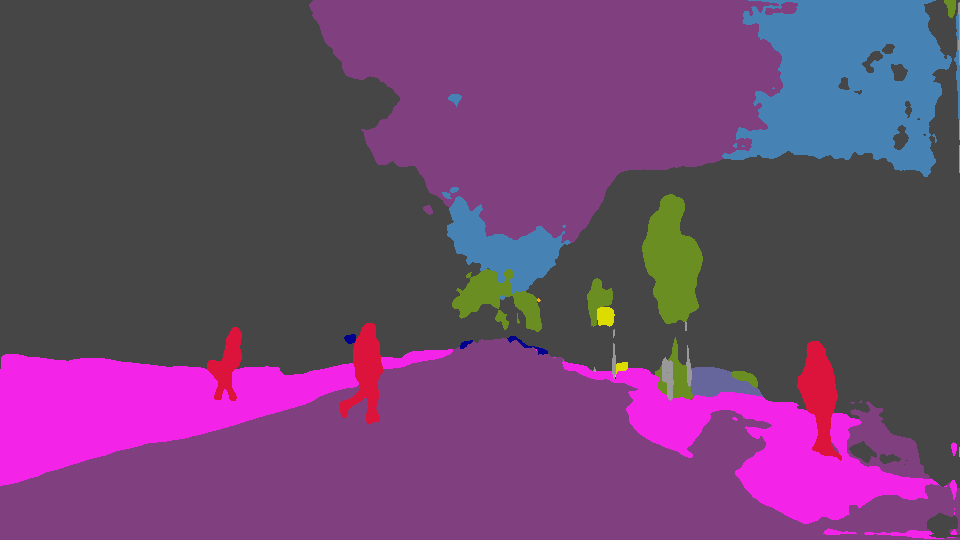} & \hspace{-.45cm}
			\includegraphics[width=.196\textwidth]{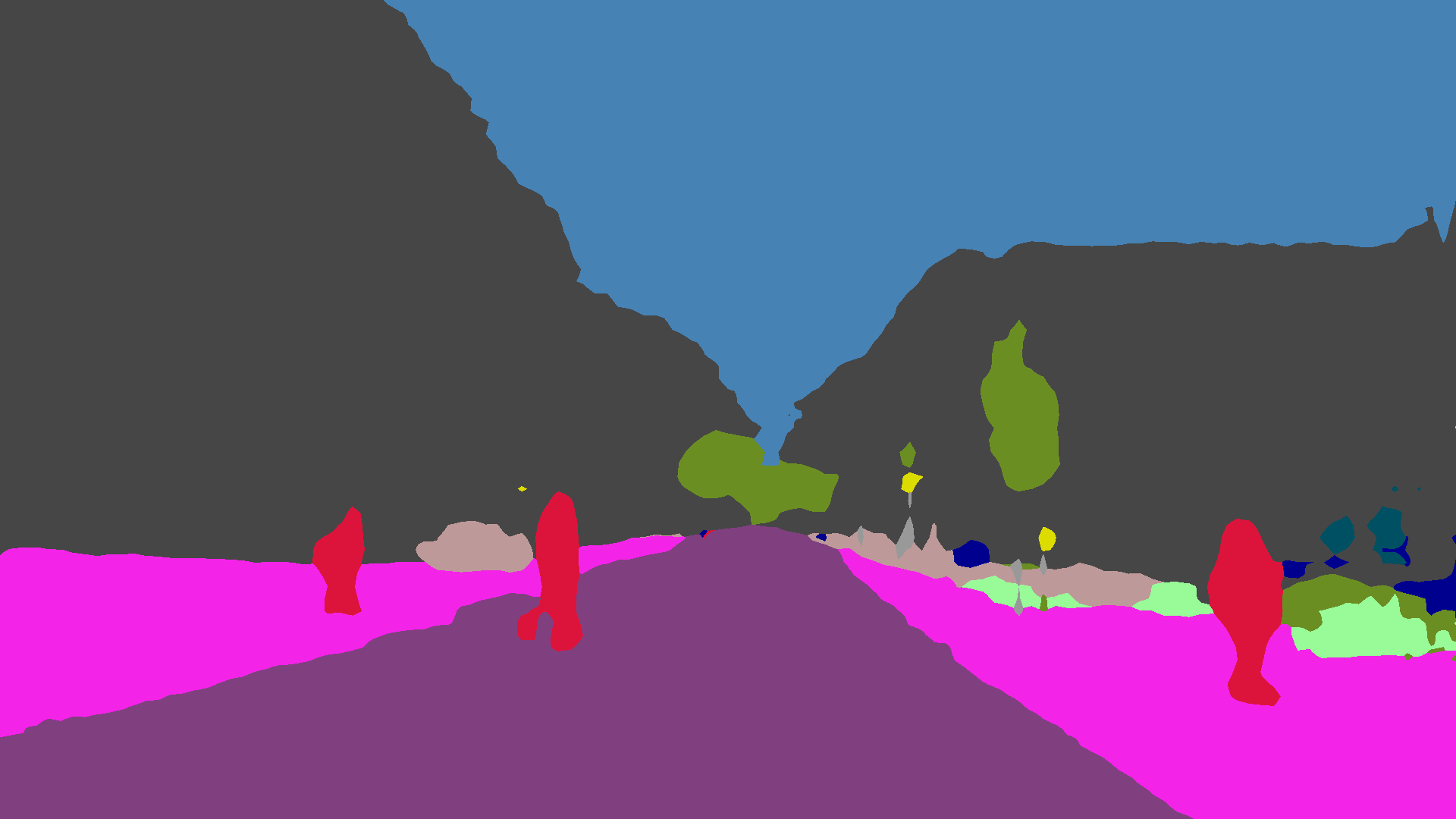} & \hspace{-.45cm}
			\includegraphics[width=.196\textwidth]{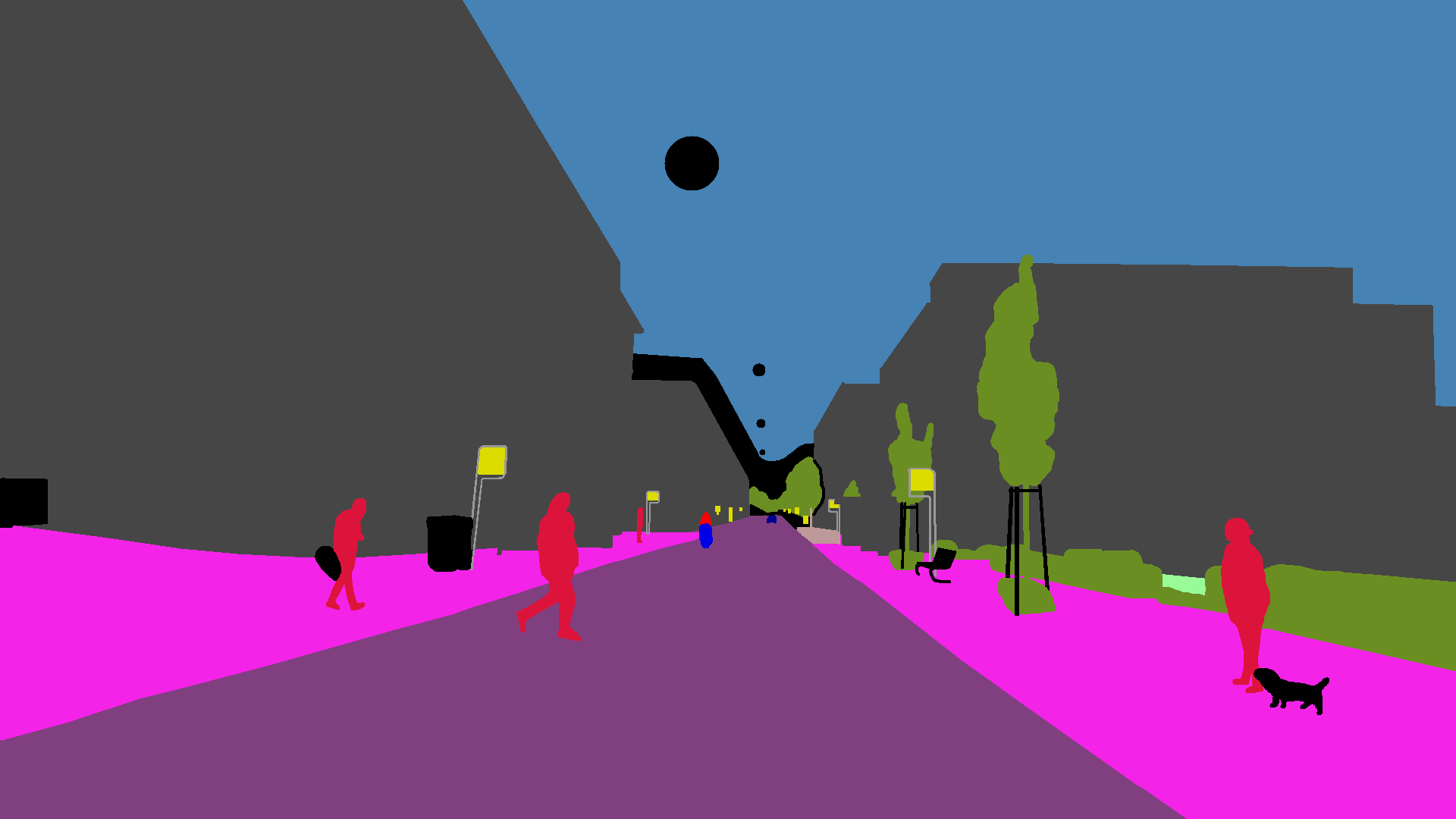} \vspace{-.05cm} \\
			\hspace{-.21cm}
			\includegraphics[width=.196\textwidth]{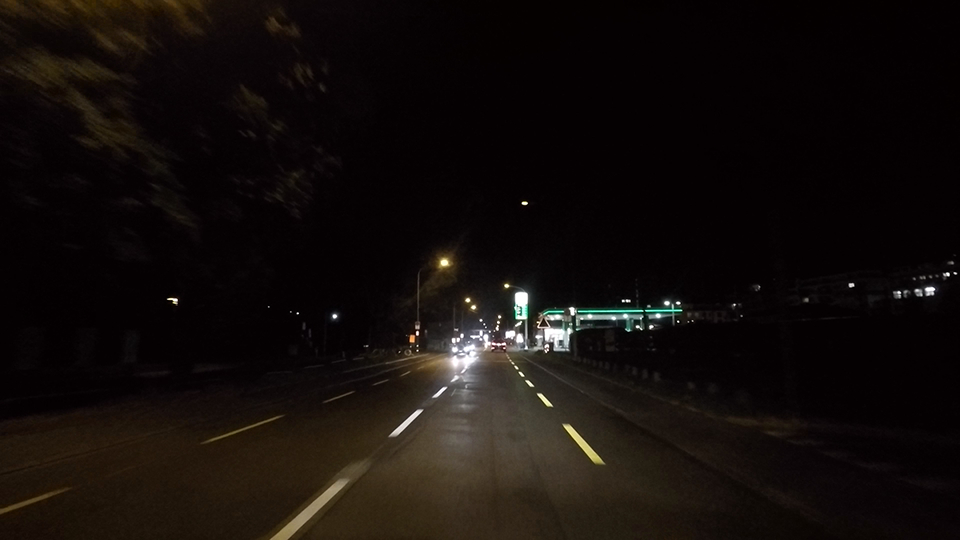} & \hspace{-.45cm}
			\includegraphics[width=.196\textwidth]{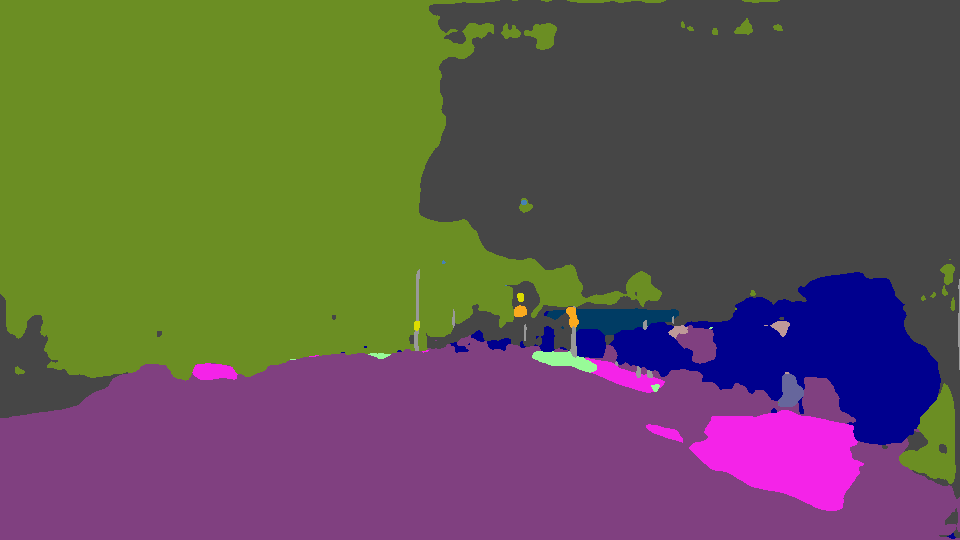} & \hspace{-.45cm}
			\includegraphics[width=.196\textwidth]{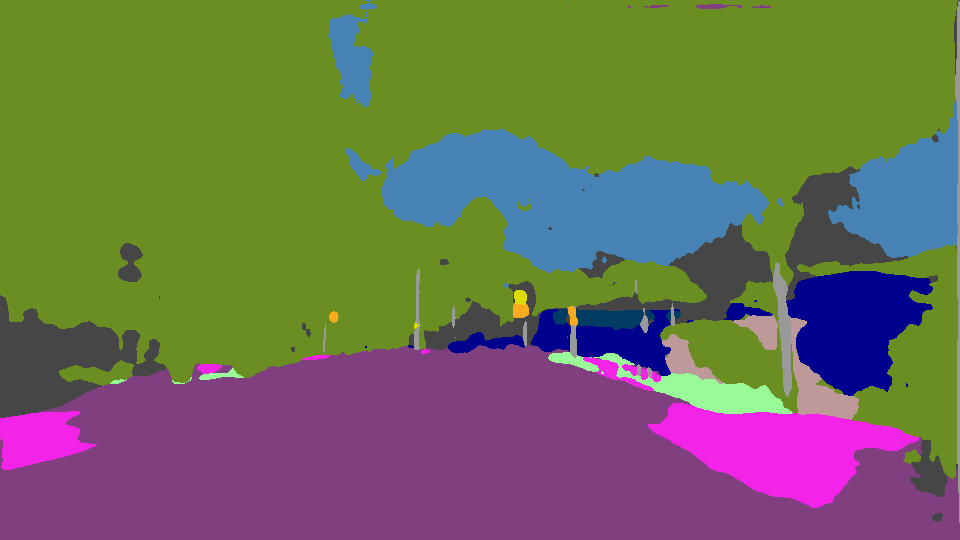} & \hspace{-.45cm}
			\includegraphics[width=.196\textwidth]{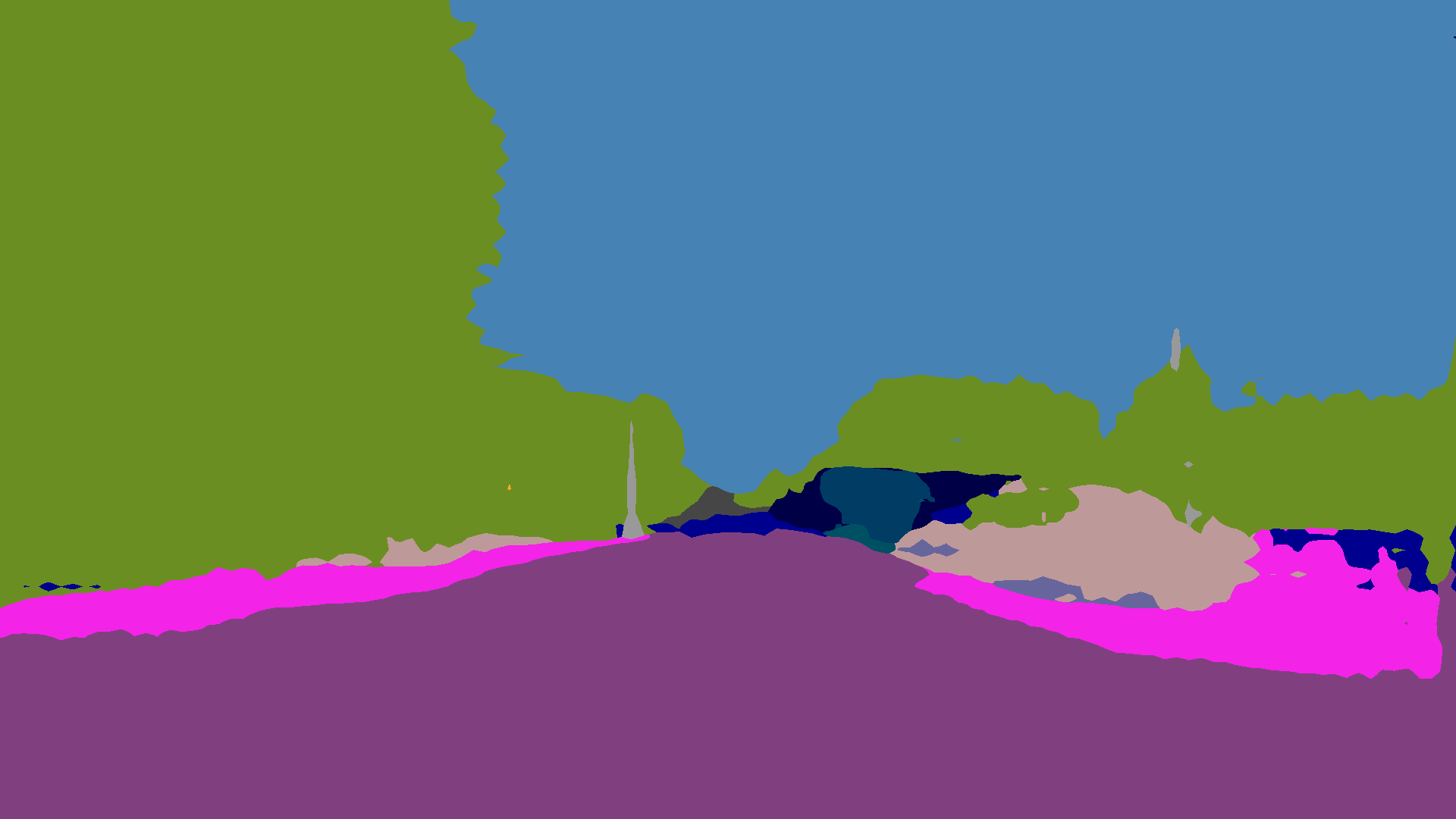} & \hspace{-.45cm}
			\includegraphics[width=.196\textwidth]{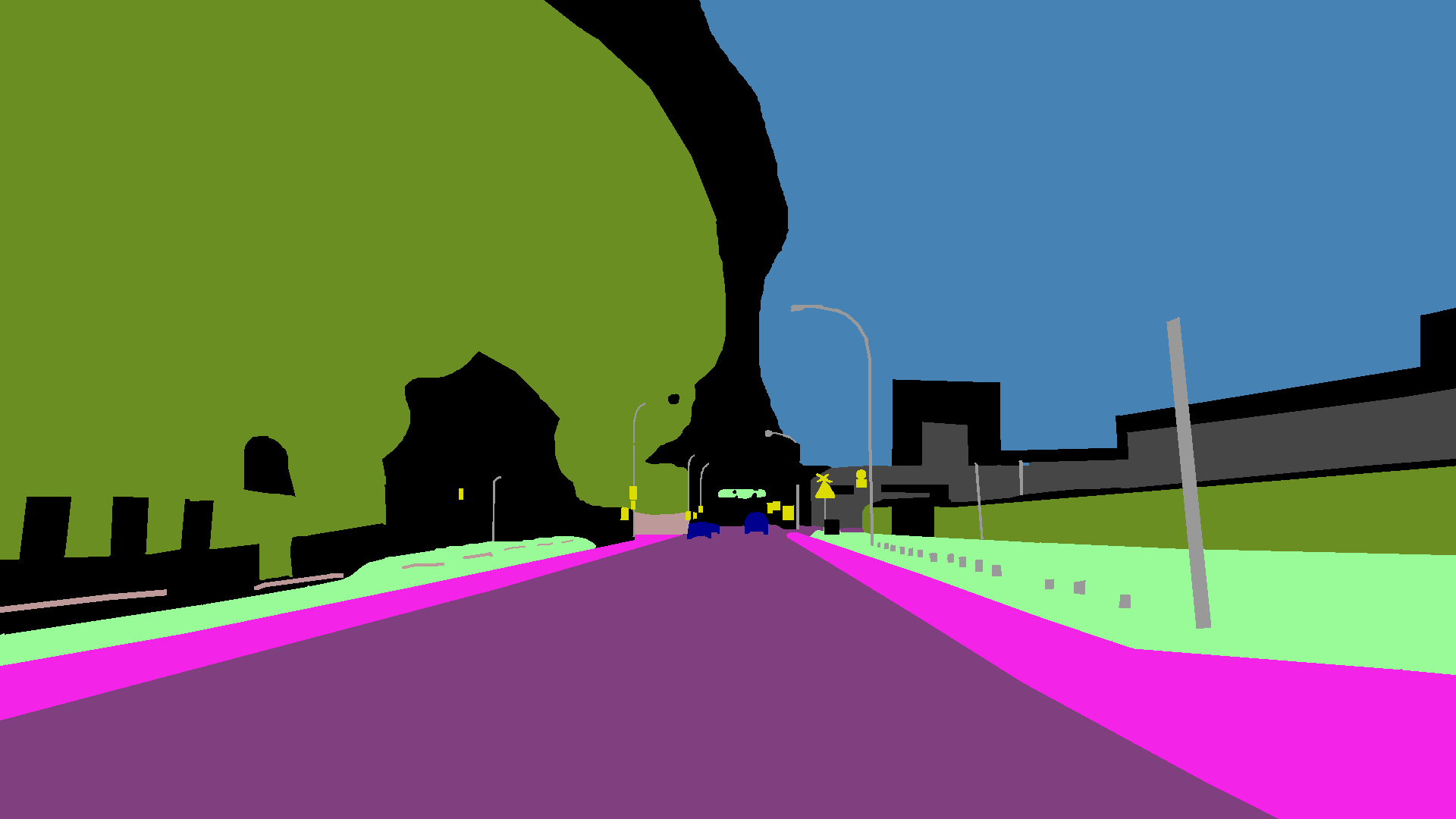} \vspace{-.05cm} \\
			\hspace{-.21cm}
			\includegraphics[width=.196\textwidth]{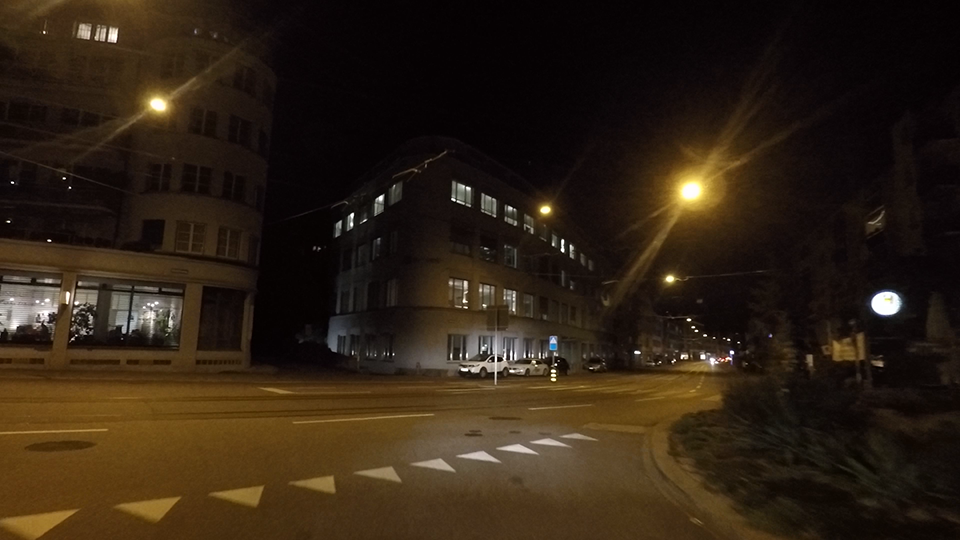} & \hspace{-.45cm}
			\includegraphics[width=.196\textwidth]{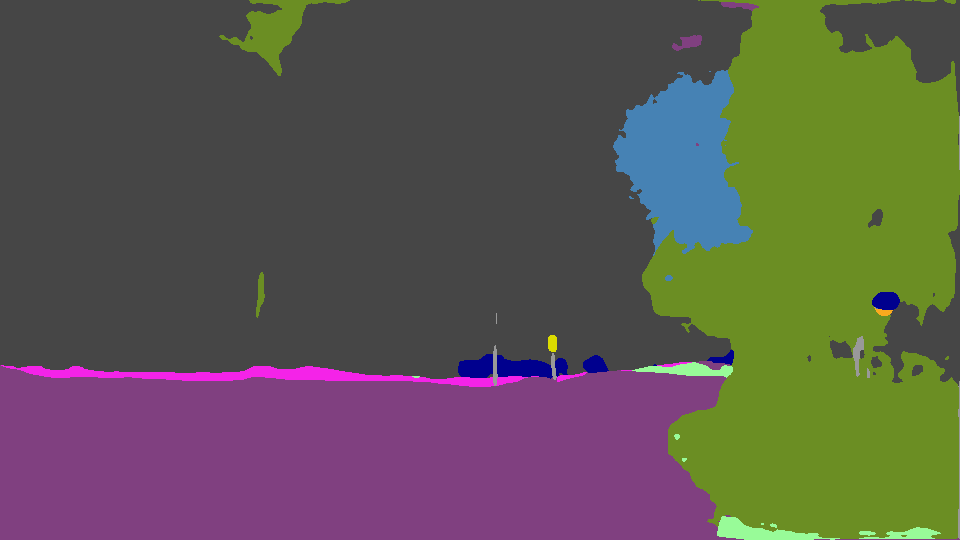} & \hspace{-.45cm}
			\includegraphics[width=.196\textwidth]{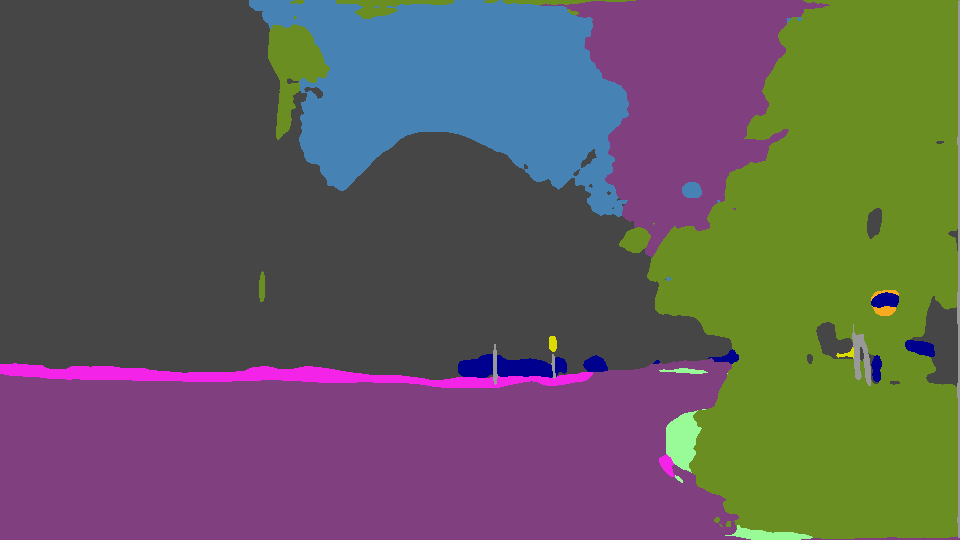} & \hspace{-.45cm}
			\includegraphics[width=.196\textwidth]{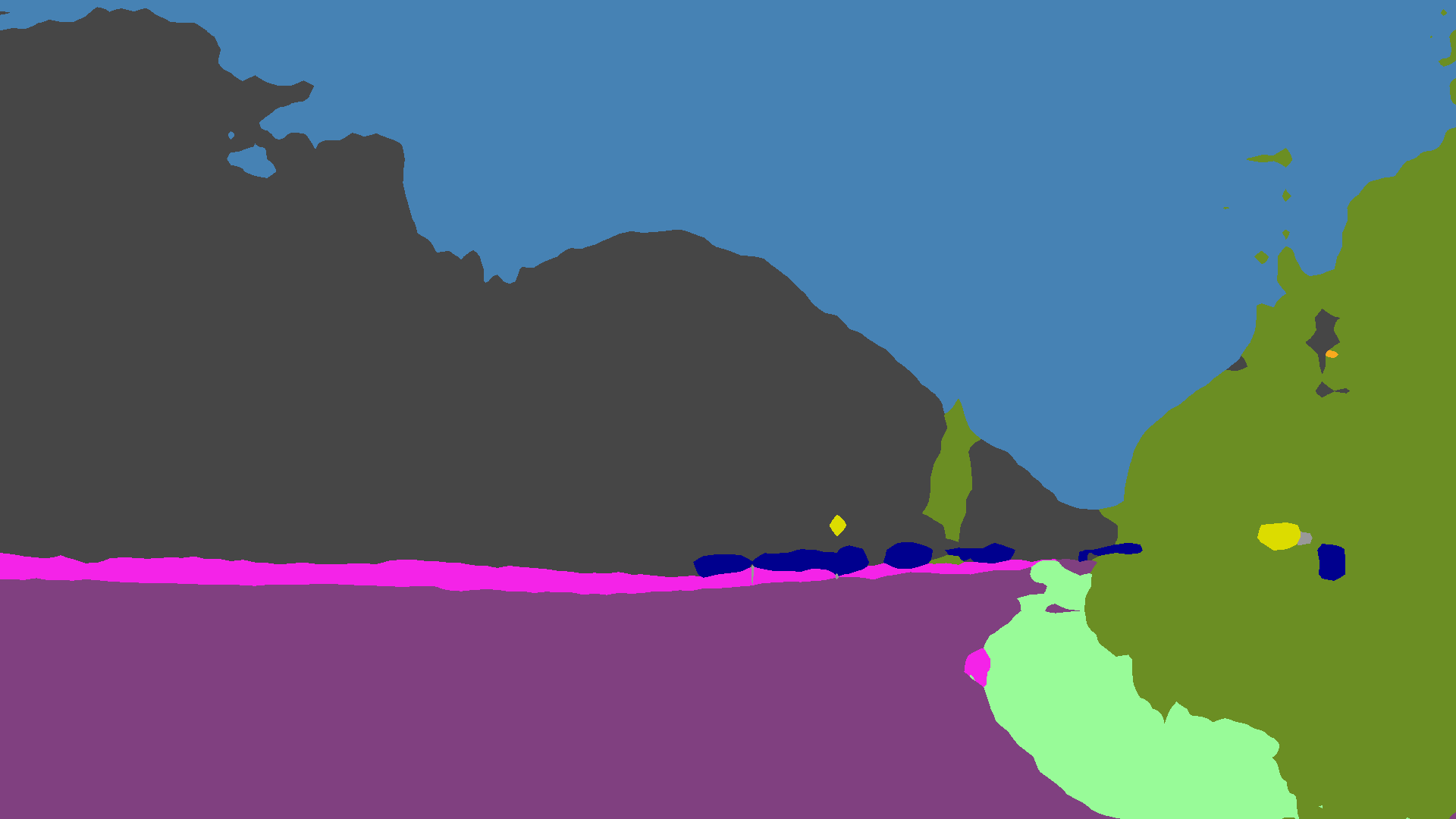} & \hspace{-.45cm}
			\includegraphics[width=.196\textwidth]{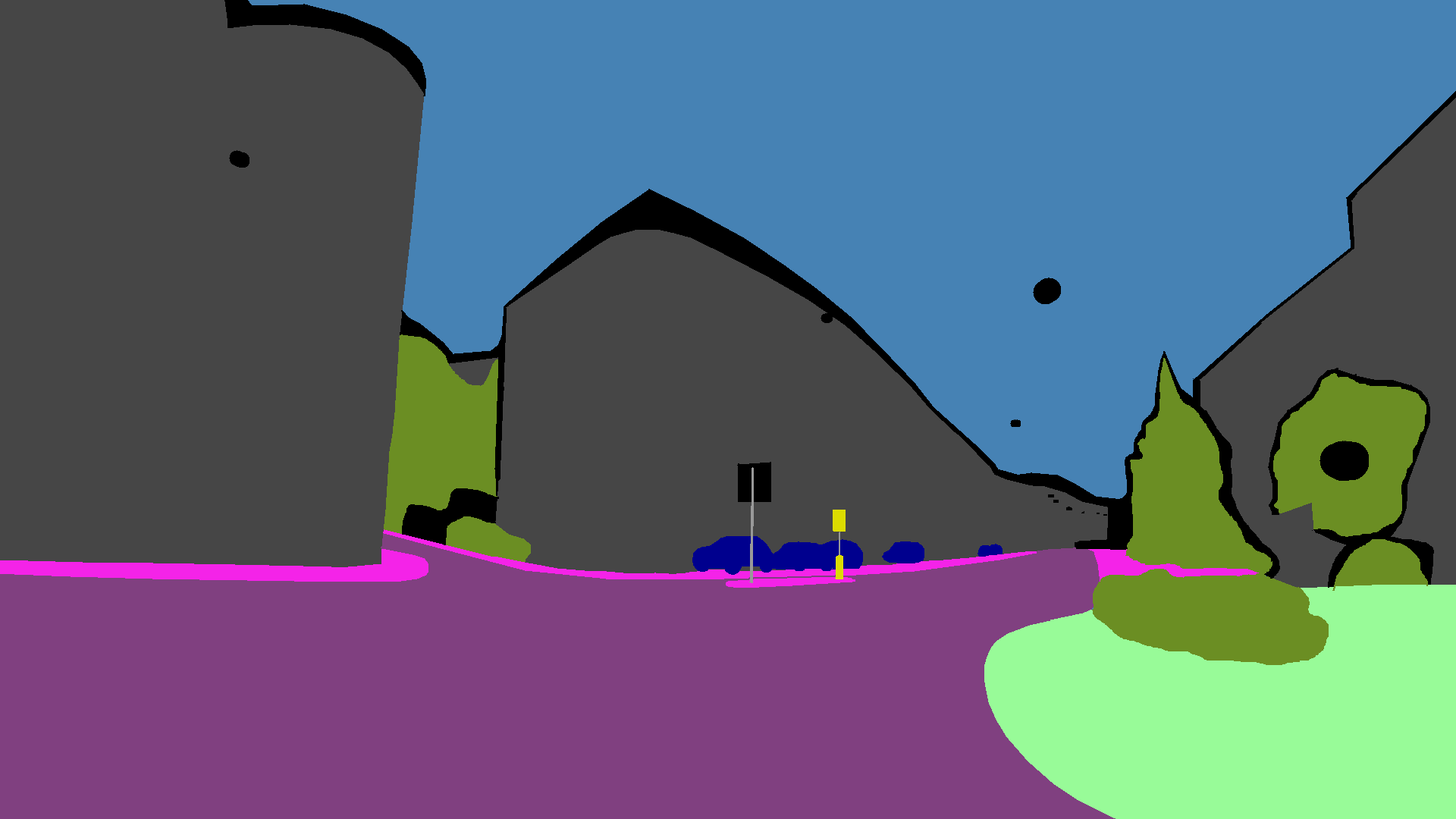} \vspace{-.05cm} \\
			\hspace{-.21cm}
			Input image   & \hspace{-.45cm}  GCMA \cite{sakaridis2019guided}  & \hspace{-.45cm} MGCDA \cite{sakaridis2020map}  & \hspace{-.45cm} DANNet (PSPNet) & \hspace{-.45cm} Semantic GT \\
		\end{tabular}
		\vspace{1pt}
		\caption{Visualization comparison of our DANNet with some existing  state-of-the-art methods on three samples from Dark Zurich-val.}
		\label{fig_comparison}
	\end{center}
	\vspace{-7pt}
\end{figure*}

\subsection{Comparison with state-of-the-art methods}

\noindent{\bf Comparison on Dark Zurich-test} \hspace{0.2cm}
We first compare our DANNet with some existing state-of-the-art methods, including MGCDA~\cite{sakaridis2020map}, GCMA~\cite{sakaridis2019guided}, DMAda~\cite{dai2018dark} and several other domain adaptation approaches~\cite{tsai2018learning,vu2019advent,li2019bidirectional} on Dark Zurich-test, and 
the results on the mIoU performance are reported in Table~\ref{DZ-test}.
Among these methods, MGCDA, GCMA, and DMAda share the same baseline RefineNet while the rest are based on Deeplab-v2 and they use the common ResNet-101 backbone~\cite{he2016deep} and the nighttime images in Dark Zurich-test as inputs during testing. 
Our DANNets with either DeepLab-v2, RefineNet or PSPNet all perform better than or tie to existing methods on this dataset, and the one with PSPNet
achieves the best performance among all, with a  2.7\% improvement of the overall mIoU over the highest  score obtained by all existing methods (by MGCDA).
We also observe that our DANNet significantly outperforms other methods on quite a few categories, such as road, sidewalk, and sky, which indicates that our method handles the large day-to-night domain gap very well even in discernible regions. 
Sample visualization results on Dark Zurich-val in Figure~\ref{fig_comparison} also verify such observation.

\begin{figure*}[h]
	\begin{center}
		\begin{tabular}{ccccc}
			\hspace{-.26cm}
			\includegraphics[width=.196\textwidth]{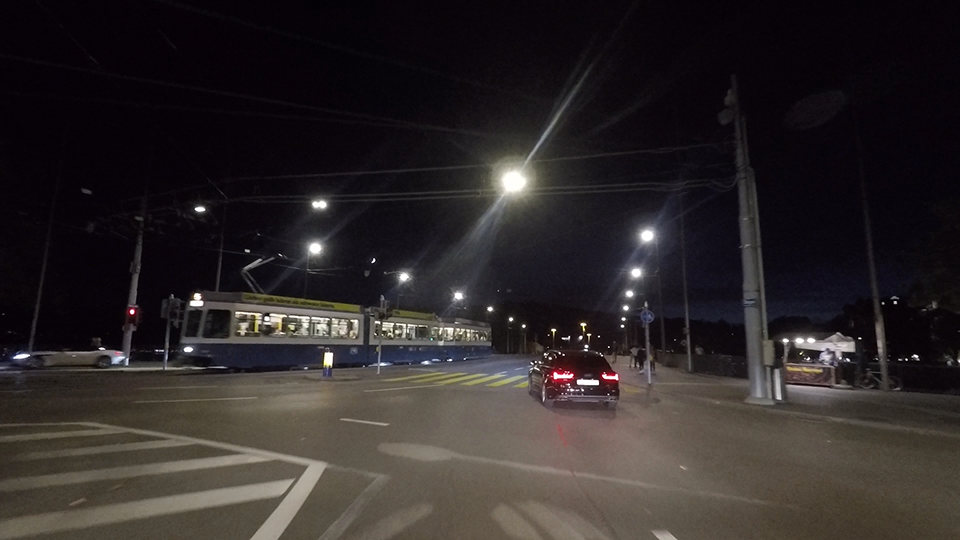} & \hspace{-.45cm}
			\includegraphics[width=.196\textwidth]{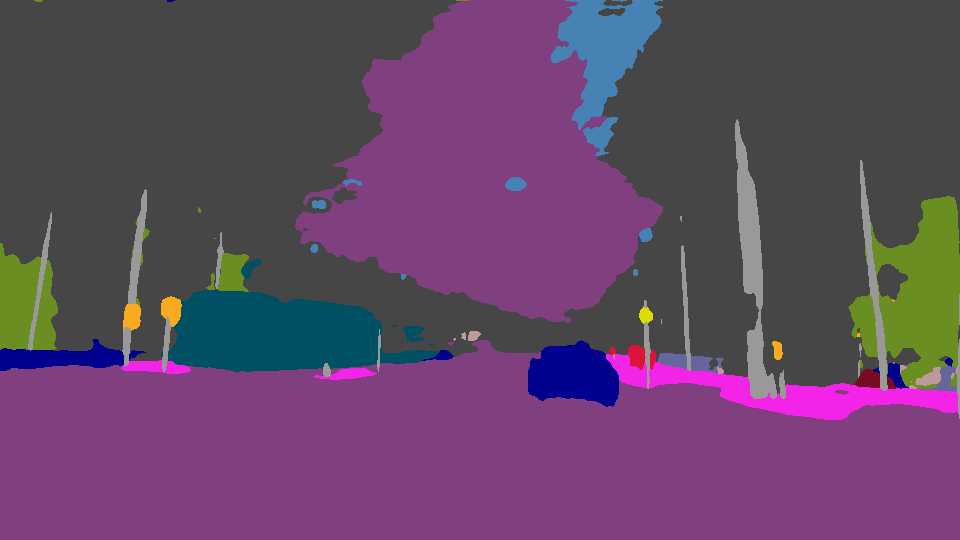} & \hspace{-.45cm}
			\includegraphics[width=.196\textwidth]{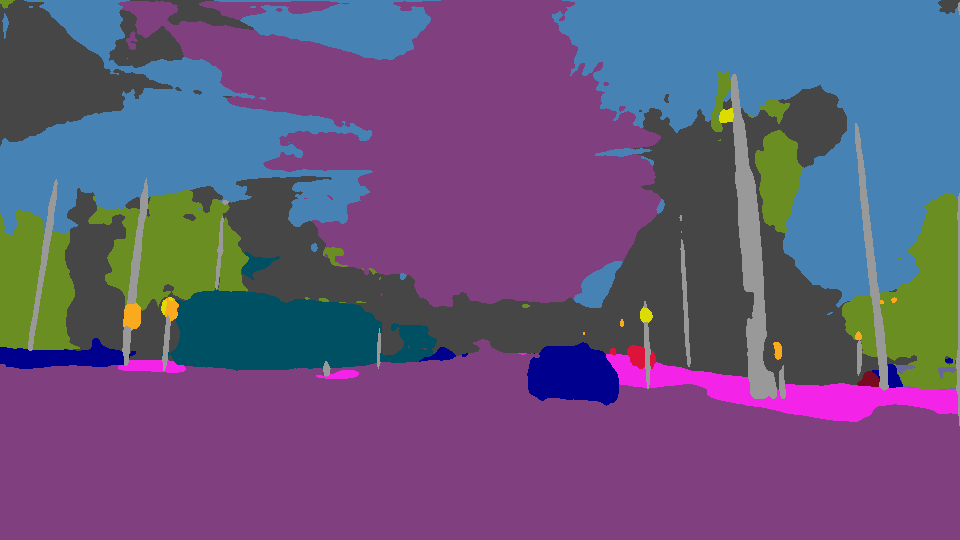} & \hspace{-.45cm}
			\includegraphics[width=.196\textwidth]{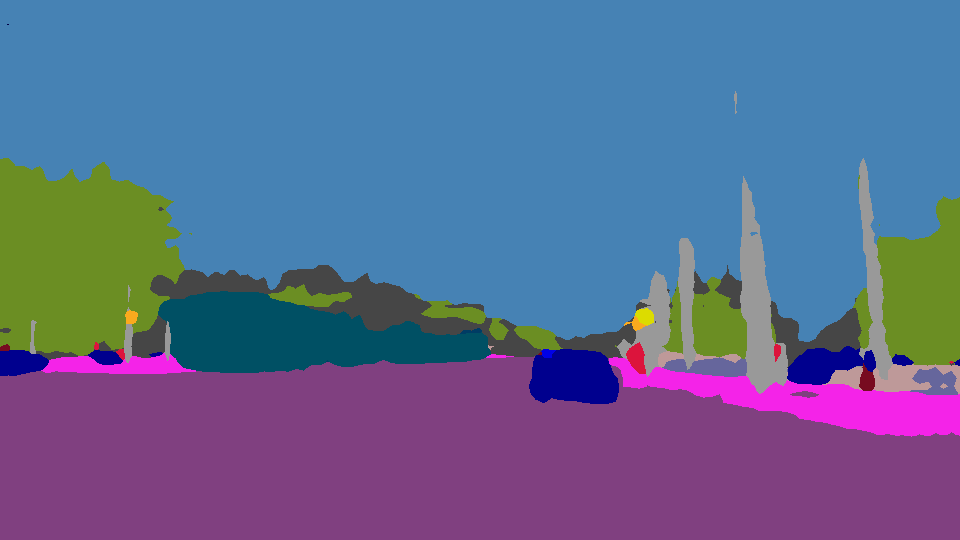} & \hspace{-.45cm}
			\includegraphics[width=.196\textwidth]{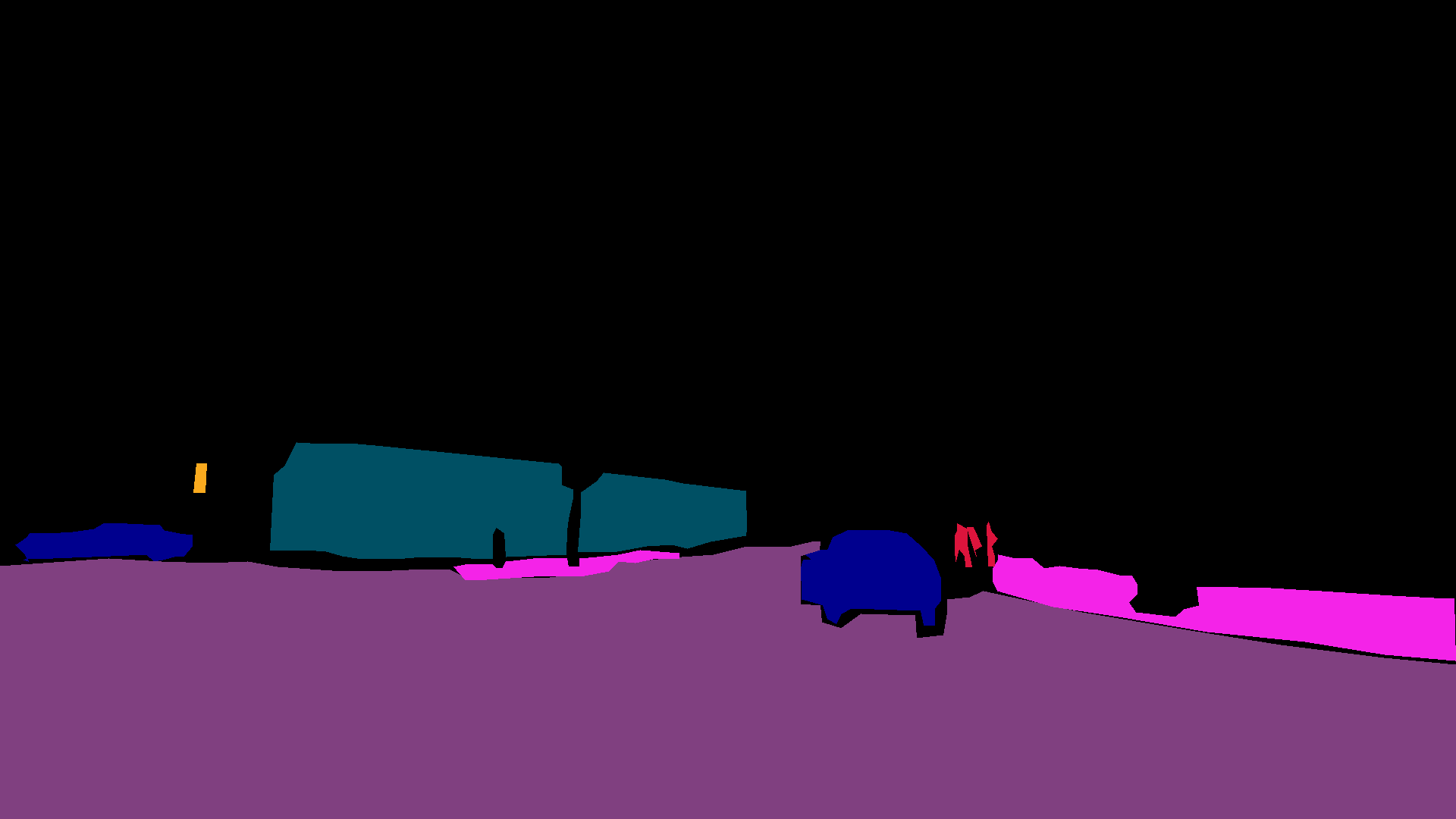} \vspace{-.05cm} \\
			\hspace{-.26cm}
			\includegraphics[width=.196\textwidth]{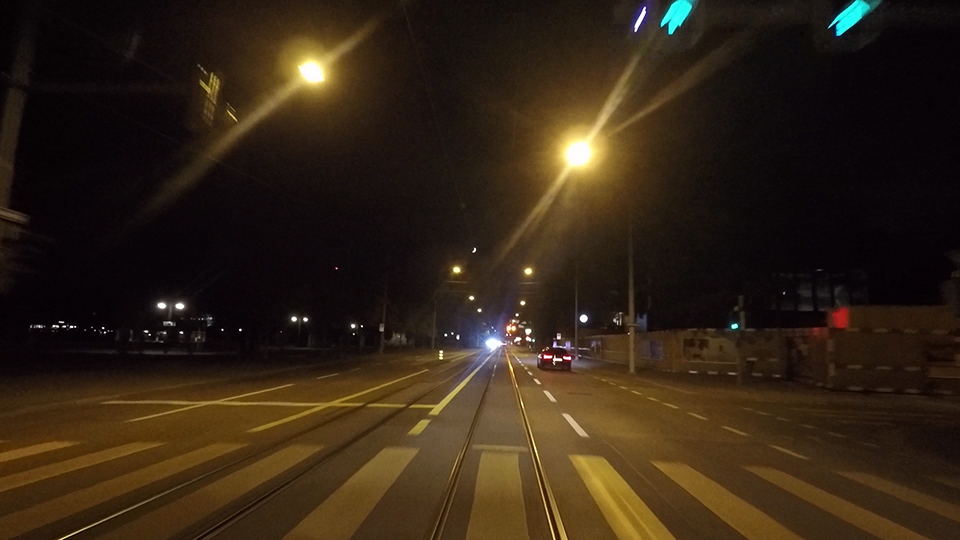} & \hspace{-.45cm}
			\includegraphics[width=.196\textwidth]{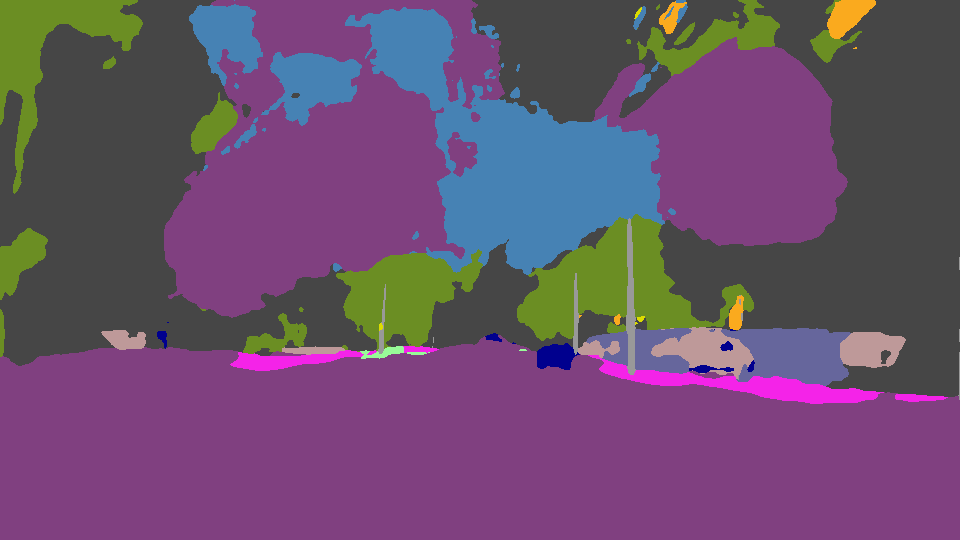} & \hspace{-.45cm}
			\includegraphics[width=.196\textwidth]{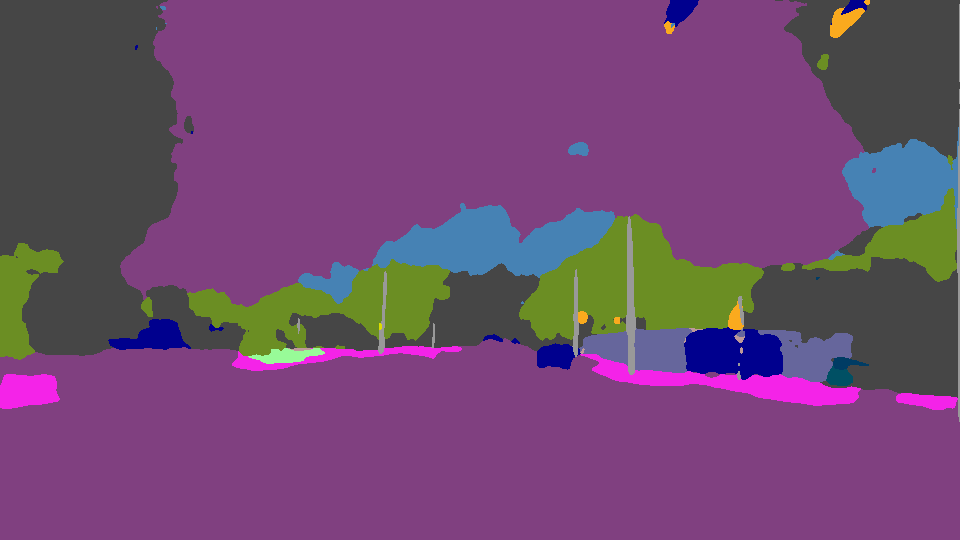} & \hspace{-.45cm}
			\includegraphics[width=.196\textwidth]{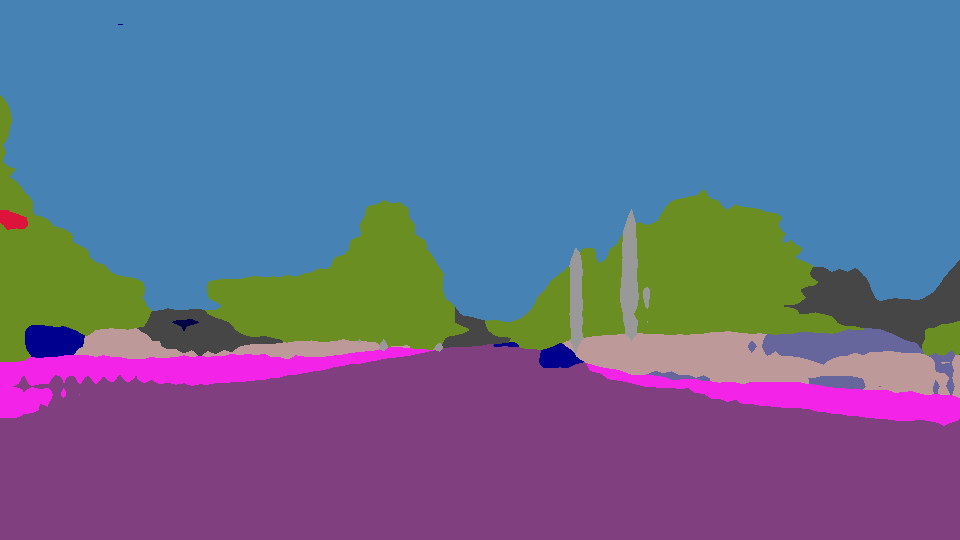} & \hspace{-.45cm}
			\includegraphics[width=.196\textwidth]{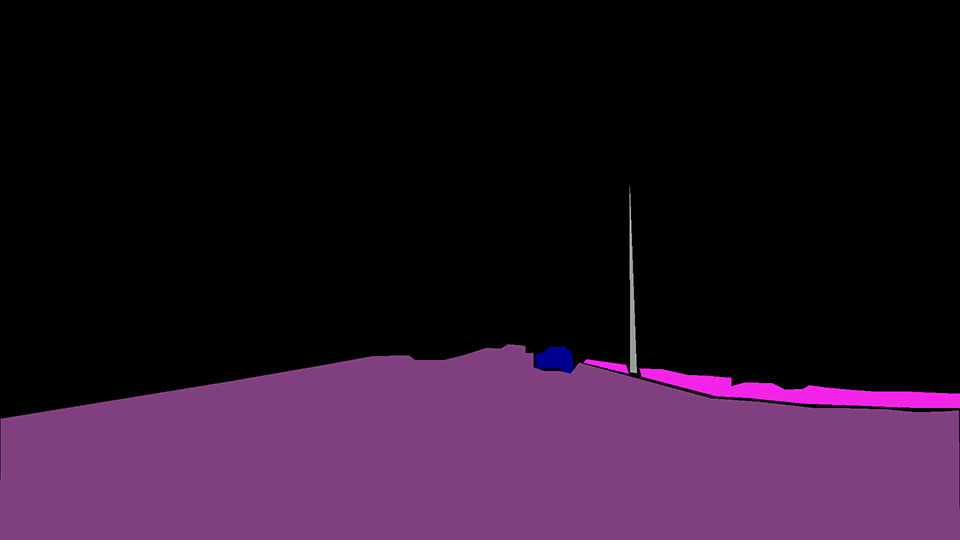} \vspace{-.05cm} \\
			\hspace{-.26cm}
			\includegraphics[width=.196\textwidth]{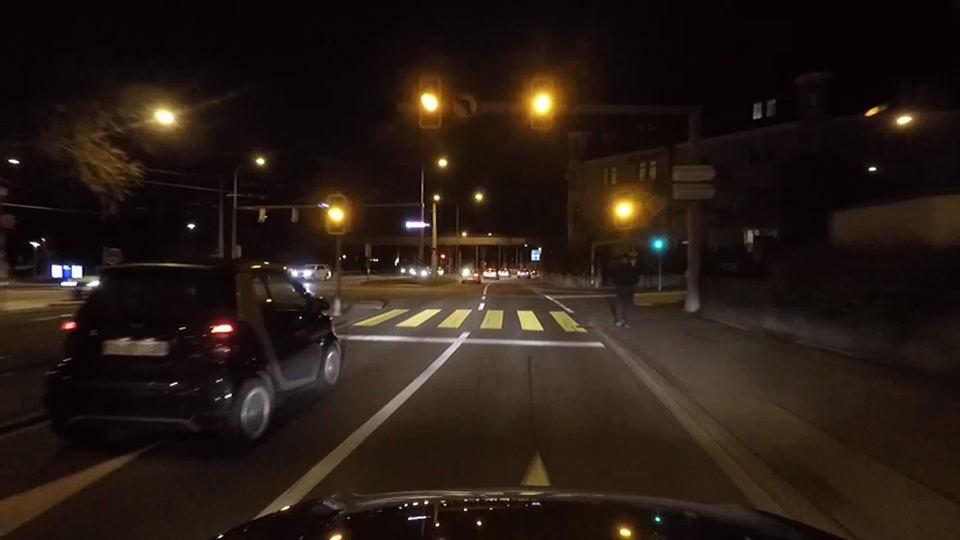} & \hspace{-.45cm}
			\includegraphics[width=.196\textwidth]{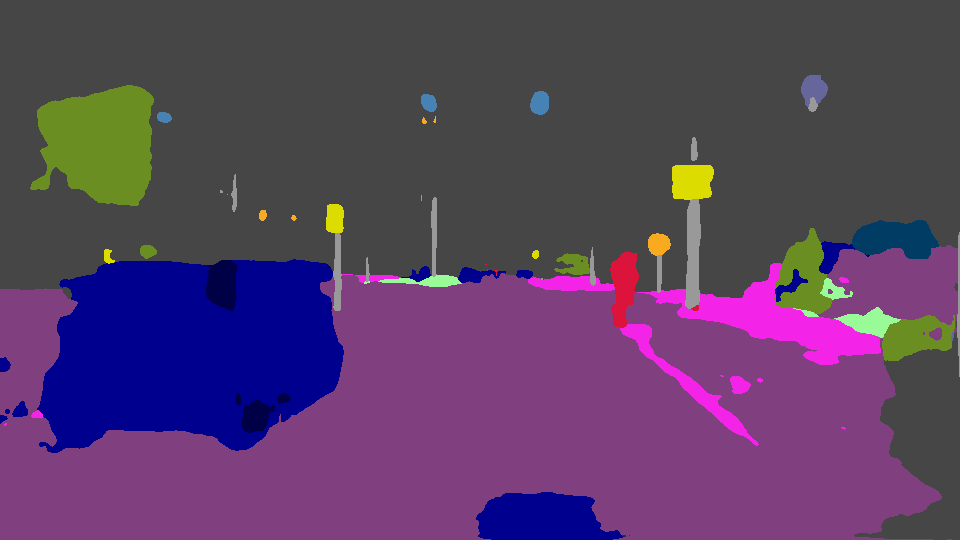} & \hspace{-.45cm}
			\includegraphics[width=.196\textwidth]{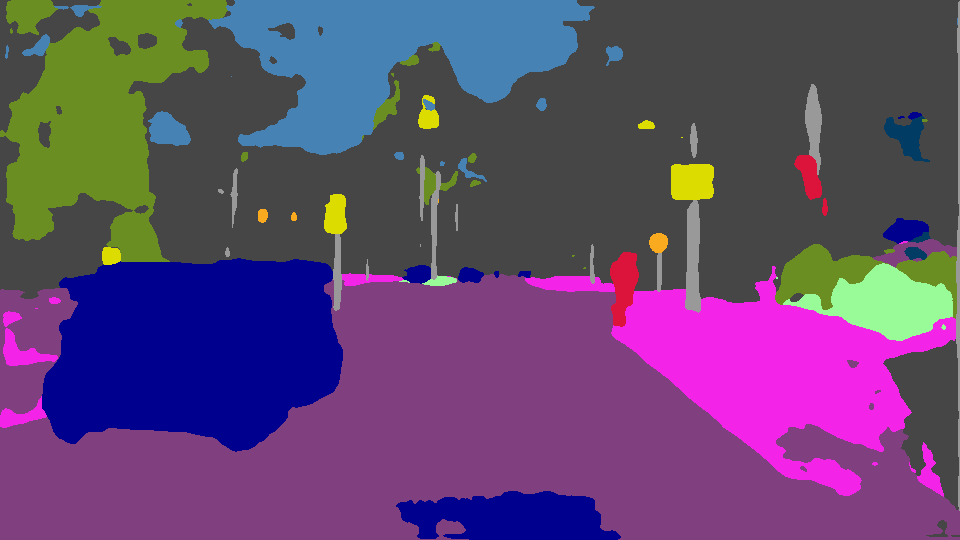} & \hspace{-.45cm}
			\includegraphics[width=.196\textwidth]{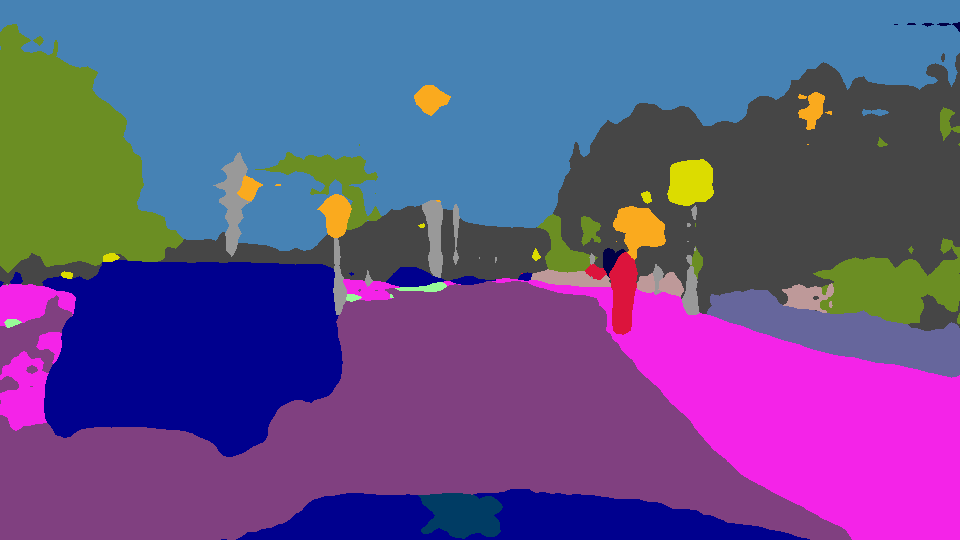} & \hspace{-.45cm}
			\includegraphics[width=.196\textwidth]{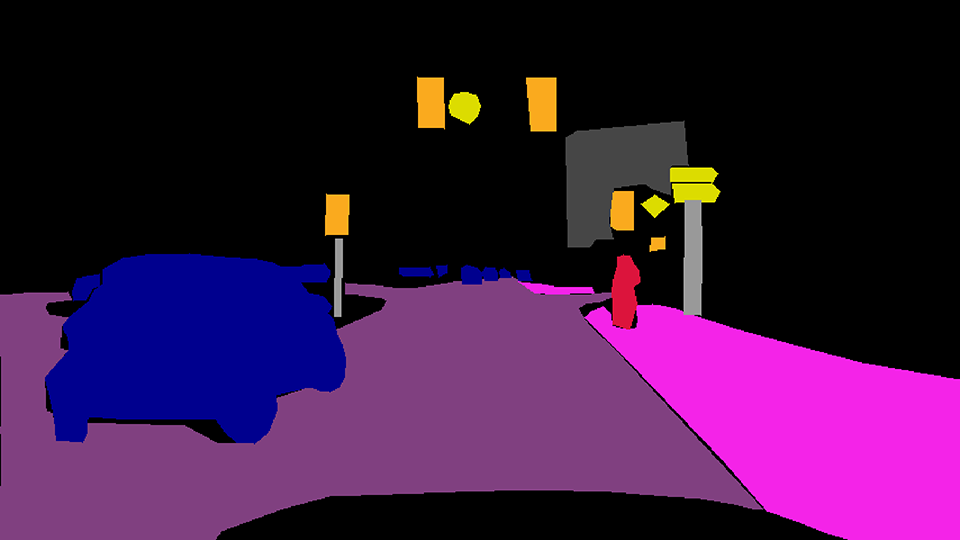} \vspace{-.05cm} \\
			\hspace{-.26cm}
			Input image   & \hspace{-.45cm}  GCMA \cite{sakaridis2019guided}  & \hspace{-.45cm} MGCDA \cite{sakaridis2020map}  & \hspace{-.45cm} DANNet (PSPNet) & \hspace{-.45cm} Semantic GT \\
		\end{tabular}
		\vspace{2pt}
		\caption{Visualization comparison of our DANNet with some existing state-of-the-art methods on three samples from Night Driving-test.}
		\label{fig_comparison_nd}
	\end{center}
	\vspace{-7pt}
\end{figure*}

\vspace{0.1cm}
\noindent{\bf Comparison on Night Driving}\hspace{0.2cm}
We report the performance of the proposed DANNet and the same set of comparison methods on Night Driving test set in Table~\ref{ND-test}, with sample 
visualization results presented in Figure~\ref{fig_comparison_nd}.
It is worth to mention that Night Driving dataset is not labeled as elaborately as Dark Zurich-test as shown in Figure~\ref{fig_comparison_nd},
and  many categories that our DANNet predicts well (see Table~\ref{DZ-test}), such as building and vegetation, are not annotated in this test set. 
We also notice that the category of sky is only labeled in 2 out of the 50 images in Night Driving test set. 
Even with these issues, our DANNet with PSPNet still achieves the second best performance (MGCDA gets the best) on this dataset. 

\begin{table}[!ht]
	\centering\footnotesize
	\caption{Comparison of our DANNet  with some existing state-of-the-art methods on Nighttime Driving test set \cite{dai2018dark}.}
	\vspace{4pt}
	\label{ND-test}
	\renewcommand\arraystretch{1.08}
	\setlength{\tabcolsep}{2mm}{
		\footnotesize
		\begin{tabular}{lc}
			\toprule
			Method  &  mIoU\\ 
			\midrule
			RefineNet \cite{Lin:2017:RefineNet}-Cityscapes  &32.75\\
			DeepLab-v2  \cite{chen2017deeplab}-Cityscapes  &25.44\\
			PSPNet \cite{zhao2017pyramid}-Cityscapes  & 27.65\\
			\midrule
			AdaptSegNet-Cityscapes$\rightarrow$DZ-night \cite{tsai2018learning}  &34.5 \\
			ADVENT-Cityscapes$\rightarrow$DZ-night \cite{vu2019advent}  &34.7\\
			BDL-Cityscapes$\rightarrow$DZ-night \cite{li2019bidirectional}  &34.7 \\
			DMAda \cite{dai2018dark}  &36.1 \\
			GCMA \cite{sakaridis2019guided}  &45.6 \\
			MGCDA \cite{sakaridis2020map} &{\bf 49.4} \\
			\midrule
			DANNet (RefineNet) & 42.36\\
			DANNet (DeepLab-v2) & 44.98\\
			DANNet (PSPNet) &\underline{47.70} \\
			\bottomrule
	\end{tabular}}
	\vspace{-5pt}
\end{table}

\subsection{Ablation study}
To demonstrate the effectiveness of different components of the proposed DANNet, we train several model variants for 35,000 epochs and test them on  Dark Zurich-val. The performance results are  reported in Table~\ref{ablation}. 
Adaptation to Dark Zurich-N using AdaptSegNet \cite{tsai2018learning} serves as the baseline and DANNet is the full model. 
We observe that coarsely aligned Dark Zurich-D is quite important although it is unlabeled, and the pseudo labels drawn from the predictions on Dark Zurich-D also play a key role in our network, without which the mIoU decreases by 13.78\%. 
Both the image relighting network and the corresponding loss $L_{light}$ can enhance the performance. 
We also see that the specially designed loss $L_{static}$ is better than directly applying the cross entropy  or  focal loss to calculate the static loss. 

\begin{table}[ht]
	\centering
	\caption{Ablation study on several model variants of our DANNet (PSPNet) on Dark Zurich-val.} 
	\vspace{4pt}
	\label{ablation}
	\renewcommand\arraystretch{1.08}
	\setlength{\tabcolsep}{2mm}{
		\small\footnotesize
		\begin{tabular}{lc}
			\toprule
			Method  &  mIoU\\ 
			\midrule
			GCMA \cite{sakaridis2019guided}  &26.65 \\
			MGCDA \cite{sakaridis2020map}  &26.10 \\
			\midrule
			AdaptSegNet-Cityscapes$\rightarrow$DZ-night \cite{tsai2018learning} &20.19\\
			\midrule
			w/o Dark Zurich-D & 22.78 \\
			w/o relighting
			 network \& $L_{light}$  & 34.14\\
			w/o $L_{light}$  & 35.05\\
			\midrule
			w/o $L_{static}$  &  20.48\\
			w/ Cross Entropy Loss in $L_{static}$  & 33.61\\
			w/ Focal Loss in $L_{static}$  & 36.49\\
			\midrule
			w/o re-weighting on pseudo labels  & 32.71\\
			w/o re-weighting on prediction  & 32.22 \\
			\midrule
			w/o pretrained segmentation model  &30.74 \\
			\midrule
			DANNet & {\bf 36.76}\\
			\bottomrule
	\end{tabular}}
	\vspace{-5pt}
\end{table}

\begin{figure}[h]\small
	\begin{center}
		\begin{tabular}{cc}
			\hspace{-.2cm}
			\includegraphics[width=.23\textwidth]{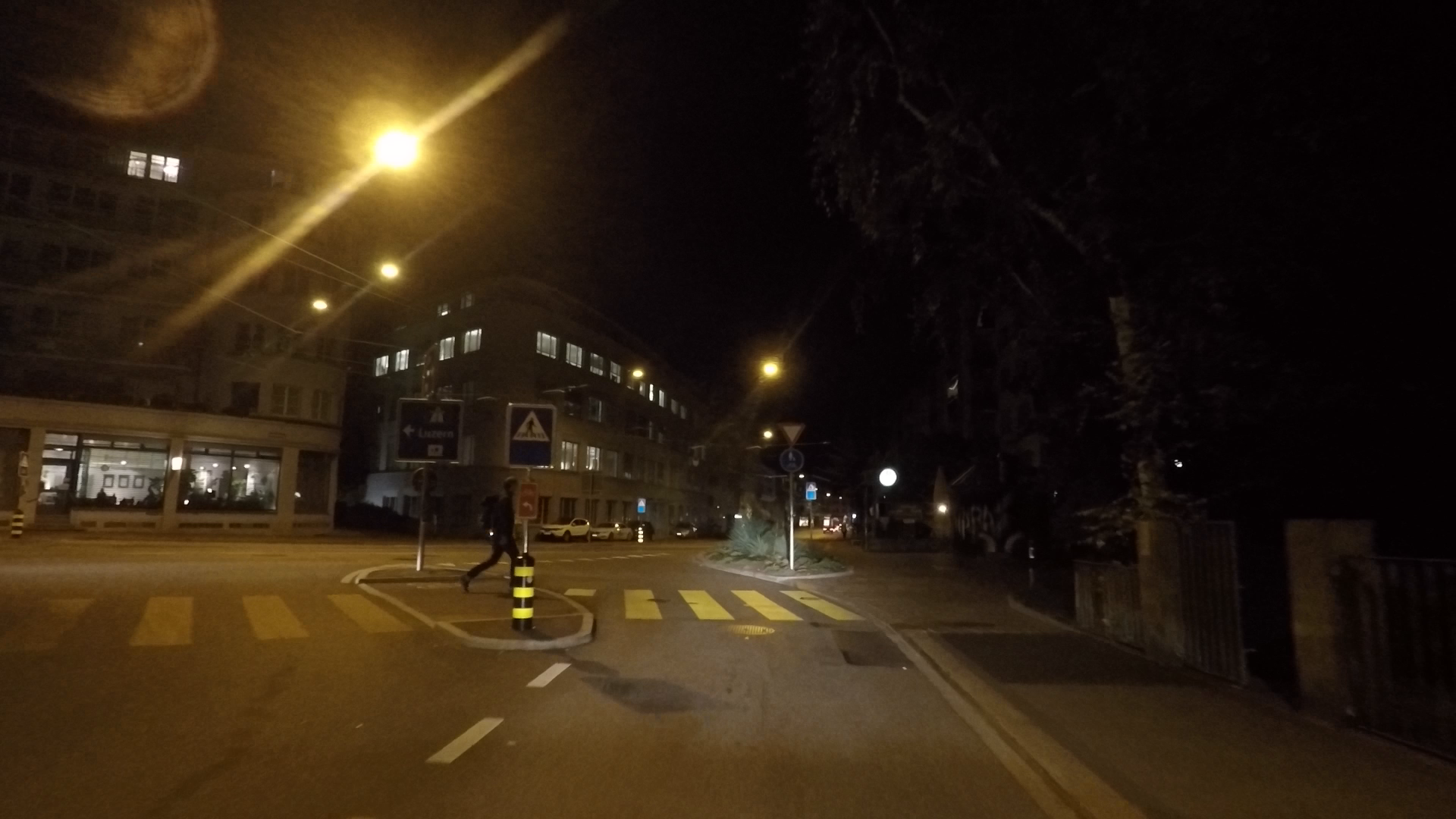} & \hspace{-.3cm}
			\includegraphics[width=.23\textwidth]{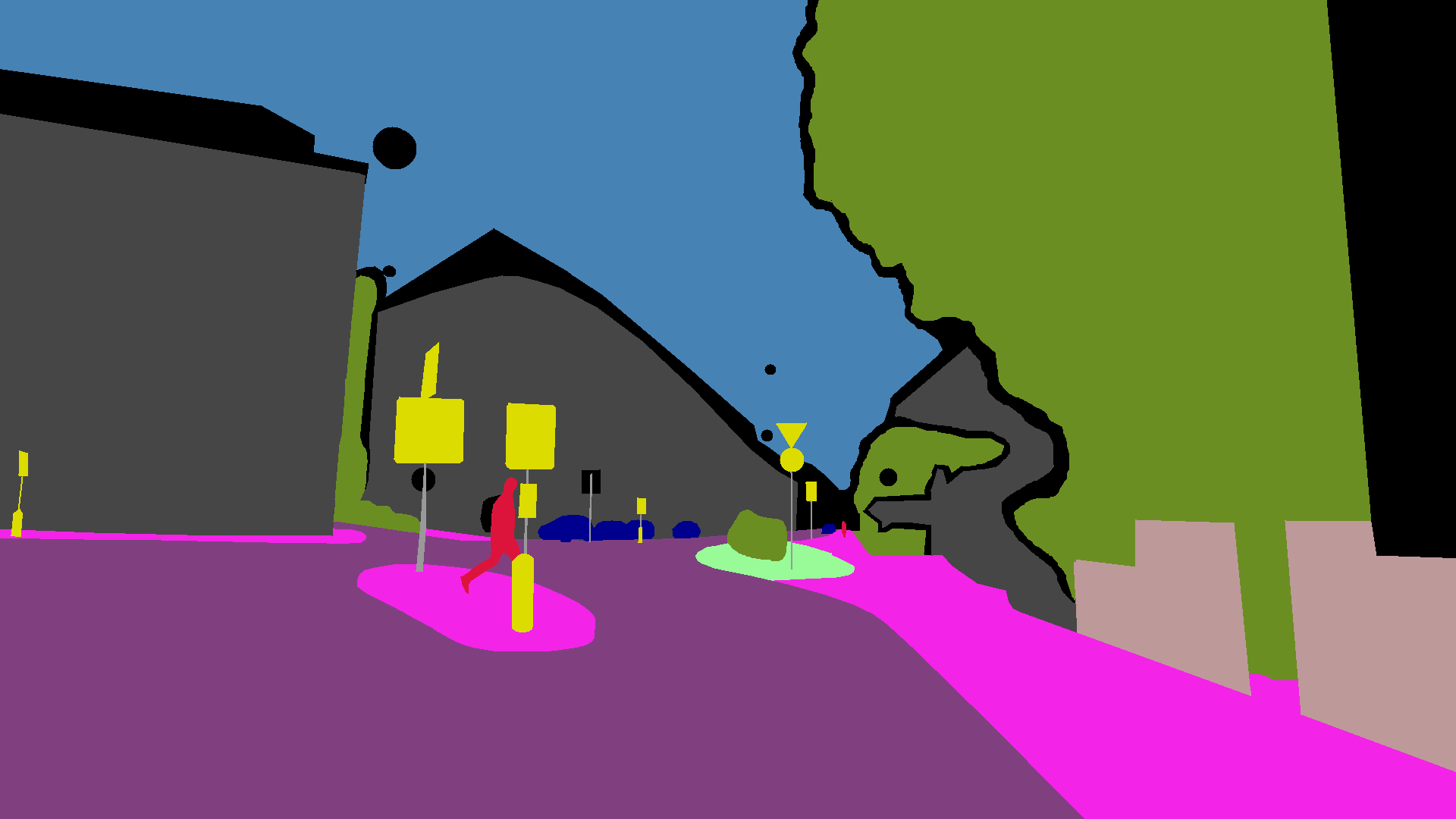} \vspace{-.05cm} \\
			Image & \hspace{-.45cm} Semantic GT   \\
			\hspace{-.2cm}
			\includegraphics[width=.23\textwidth]{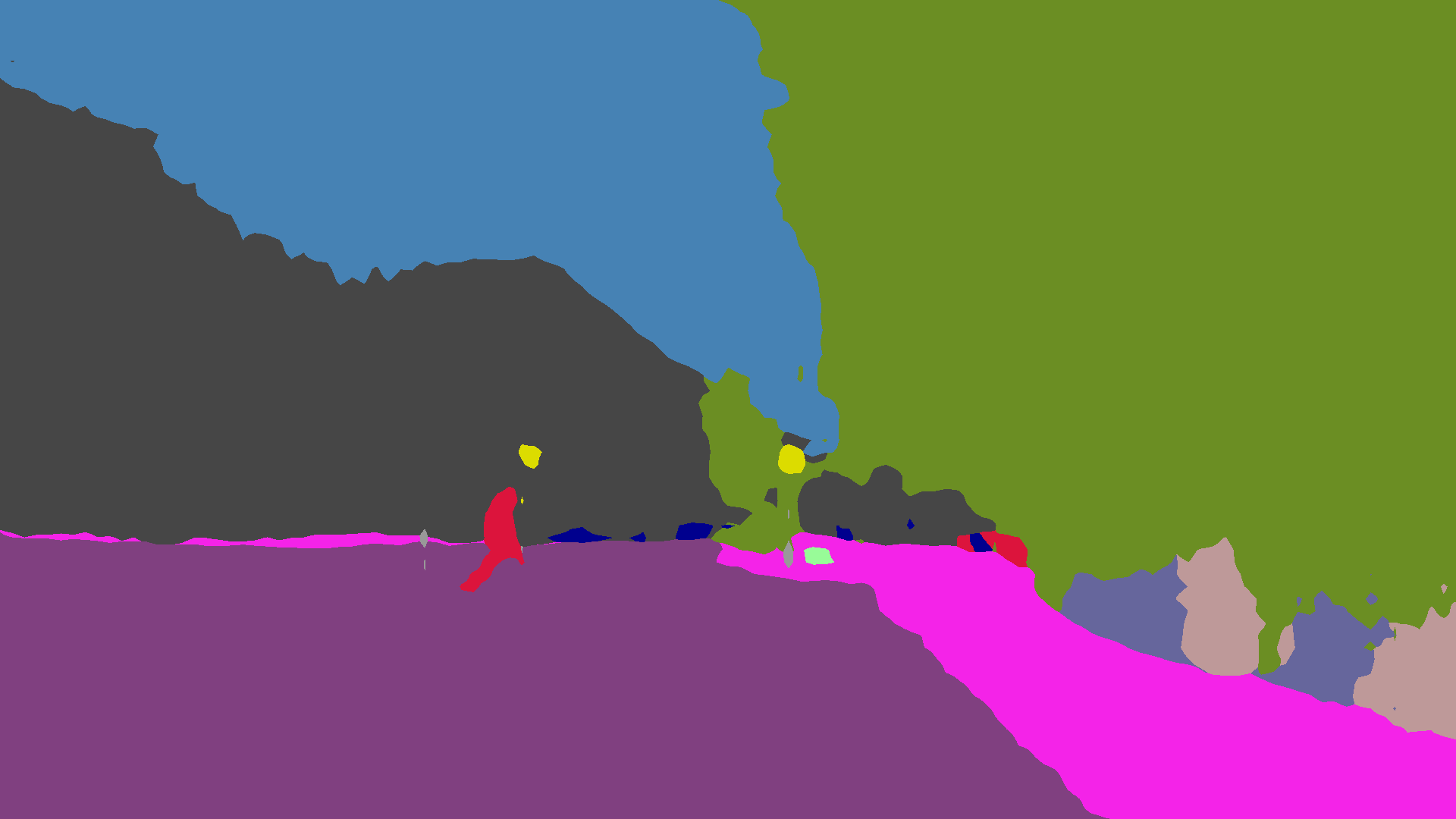} & \hspace{-.3cm}
			\includegraphics[width=.23\textwidth]{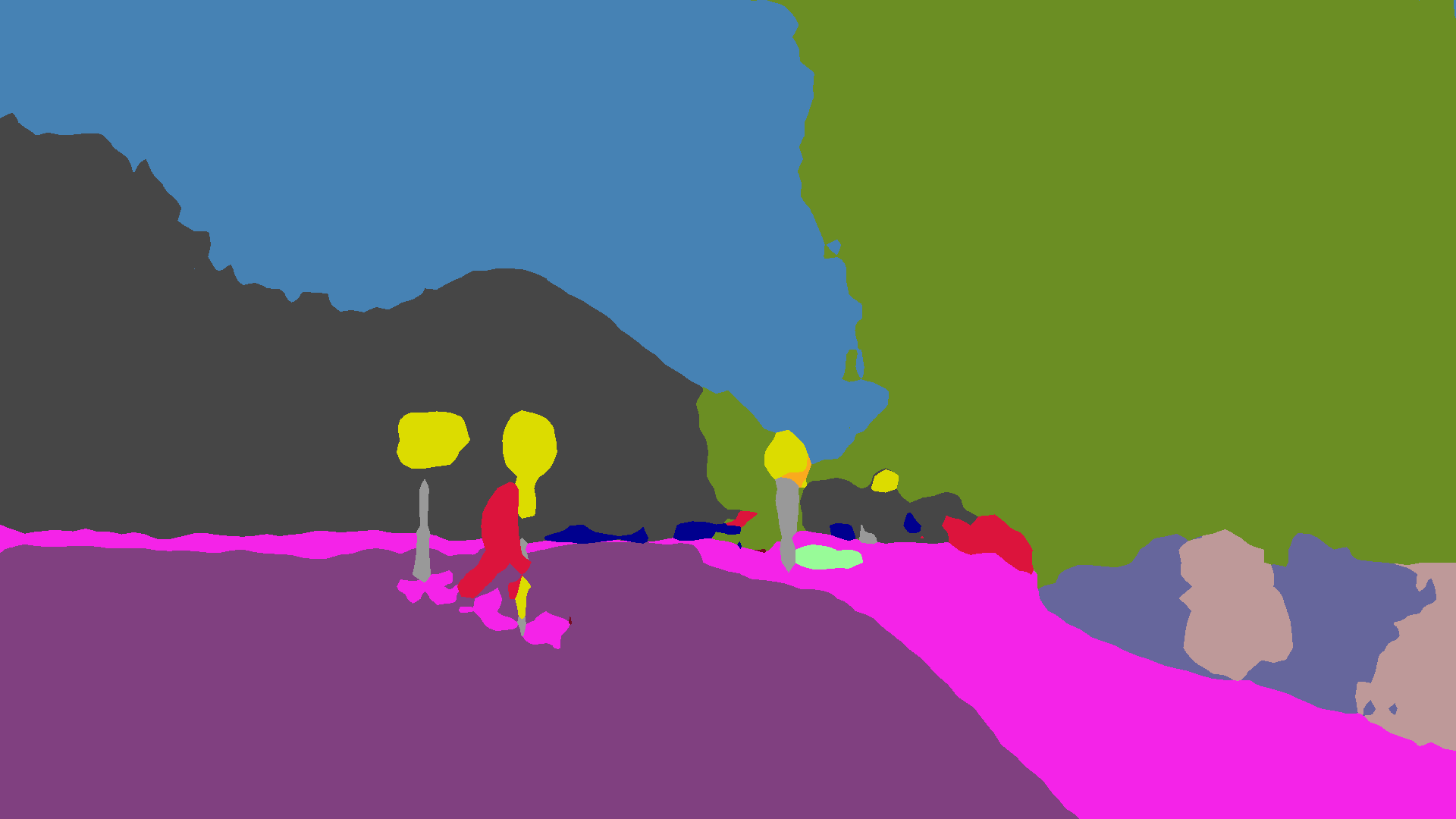} \vspace{-.05cm} \\
			\hspace{-.2cm}
			w/o re-weighting & \hspace{-.45cm} w/ re-weighting   \\
		\end{tabular}
		\vspace{1pt}
		\caption{Visualization results of w/ and w/o the re-weighting strategy on a sample from Dark Zurich-val by our DANNet (PSPNet).}
		\label{fig_ablation}
	\end{center}
	\vspace{-8pt}
\end{figure}

\begin{figure}[h]\small
	\begin{center}
		\includegraphics[width=.47\textwidth]{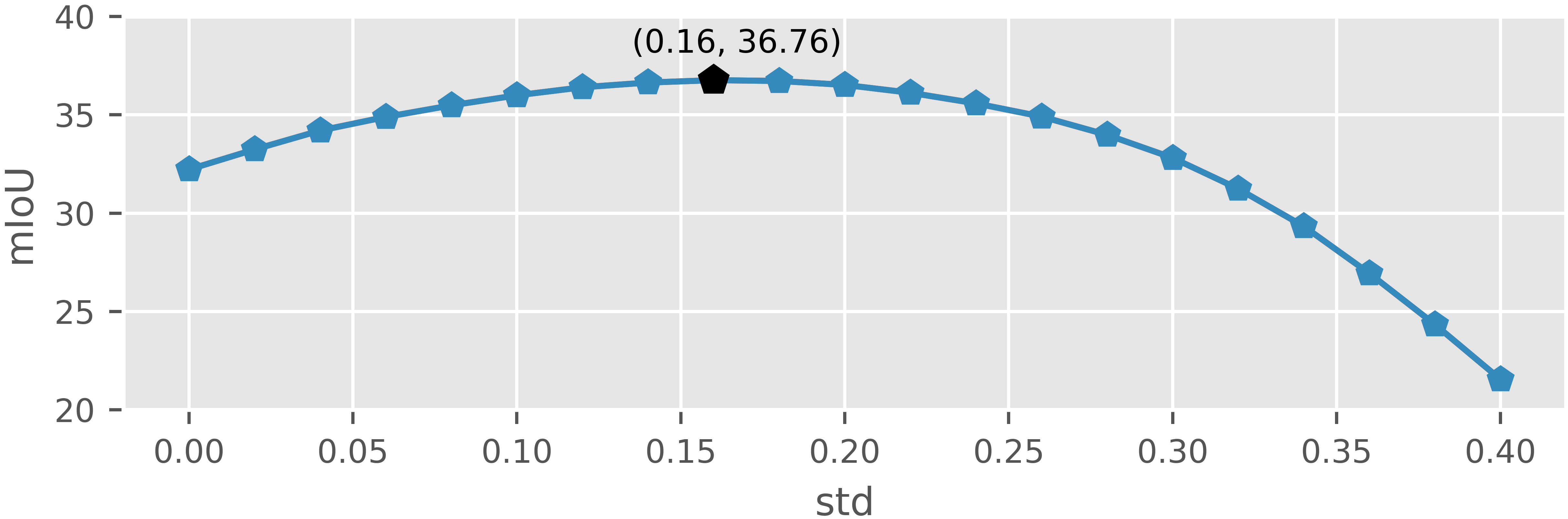}
		\vspace{1pt}
		\caption{Ablation study on the value of $std$ in the re-weighting strategy on Dark Zurich-val by our DANNet (PSPNet).}
		\label{rw_ablation}
	\end{center}
	\vspace{-18pt}
\end{figure}

In addition, the re-weighting strategy is verified to be useful and can further boost the performance. 
As shown in Figure \ref{fig_ablation}, this strategy helps segment the small objects.  
We find that the selection of the value $std$ is also important in applying the re-weighting strategy.
We test different $std$ values and the performance curve of the proposed DANNet on Dark Zurich-val is shown in Figure~\ref{rw_ablation}, and the quantitative comparison result for each category is provided in the supplemental material. 
The optimal performance is achieved when setting $std=0.16$ during testing.   By directly applying the commonly-used weights provided by OCNet~\cite{YuanW18}, it only achieves 35.05 mIoU on DZ-val dataset, which is less than that of our DANNet.
In general, the full settings of our DANNet bring about an additional 10\% performance increase over the state-of-the-art approaches on Dark Zurich-val. 

\section{Conclusion}
In this paper, we have proposed a novel end-to-end neural network DANNet for unsupervised nighttime semantic segmentation, which performs an adaptation from a labeled daytime dataset to unlabeled day-night image pairs. 
In our DANNet, an image relighting network with a special light loss function is first used to make the intensity distributions of the images from different domains to be close to each other. Then the unlabeled Dark Zurich-D data is used to bridge the domain gap between the labeled daytime images (Cityscapes) and the unlabeled nighttime images (Dark Zurich-N). By leveraging the similar illumination patterns between Dark Zurich-D and Cityscapes and coarse alignment of static categories between Dark Zurich-D and Dark Zurich-N, our DANNet performs multi-target domain adaptation as well as a re-weighting strategy to boost the performance for small objects. Experimental results demonstrated  the effectiveness of each of the designed components and showed that our DANNet achieves the state-of-the-art performance on Dark-Zurich and Night Driving test datasets. 

{\small
	\bibliographystyle{ieee_fullname}
	\bibliography{cvpr2080}
}

\end{document}